%% file: neurips_2026.tex
\documentclass{article}

\PassOptionsToPackage{numbers, compress}{natbib}

\usepackage[eandd, final]{neurips_2026}

\usepackage[utf8]{inputenc} 
\usepackage[T1]{fontenc}    
\usepackage{hyperref}       
\usepackage{url}            
\usepackage{booktabs}       
\usepackage{amsfonts}       
\usepackage{nicefrac}       
\usepackage{microtype}      
\usepackage{xcolor}         
\usepackage[table]{xcolor}
\usepackage{multirow}
\usepackage{makecell}
\usepackage{pifont}
\usepackage{graphicx}
\usepackage{amsmath}
\usepackage{amssymb}
\usepackage[breakable]{tcolorbox}
\tcbuselibrary{skins, breakable}
\usepackage{caption}
\usepackage{newunicodechar}
\usepackage{wrapfig} 
\usepackage{enumitem}

\title{AgentHop: A Diagnostic Benchmark for Agentic Multi-Hop Question Answering over Scientific Citation Graphs}

\author{
  \textbf{Chanhee Park}$^{1}$,
  Jeongho Yoon$^{1}$,
  Sungbin Han$^{1}$,
  Hyeonseok Moon$^{2}$\thanks{Corresponding authors.},
  Heuiseok Lim$^{1}$\footnotemark[1] \\
  $^{1}$Korea University \quad $^{2}$Sookmyung Women's University \\
  \texttt{\{pch7678,aa007878,sungbinhan9039,limhseok\}@korea.ac.kr} \\
  \texttt{hyns.moon@sookmyung.ac.kr}
}

\begin{document}

\maketitle

\begin{abstract}
Agentic tasks require a large language model to interact with the world, navigating information and gathering evidence across multiple steps with restricted resources. Due to this complexity, agentic task failures arise from various sources, and pinpointing these failure causes is essential to diagnose and improve agentic systems. Existing benchmarks, however, tend to focus on a single leaderboard score, leaving the underlying failure modes opaque. To fill this gap, we introduce \textbf{AgentHop}, a diagnostic benchmark of $1{,}011$ multiple-choice questions paired with a controlled seven-tool sandbox under fixed token, turn, and tool-call constraints. AgentHop reveals model vulnerabilities by dissecting a single accuracy score along four axes of agent operation: retrieval, synthesis, tool-call, and resource management. Across $19$ models, we find that behavior clusters by model family, with tool-call signatures revealing distinct family fingerprints: GPT models commit early, Anthropic and GLM checkpoints verify before committing, DeepSeek and Kimi over-search, and Gemini-3 Pro stays balanced. Decomposed axes further expose within-family structure: Claude Opus 4.6 and Sonnet 4.6 land within one accuracy point yet diverge on retrieval-versus-synthesis emphasis, with Opus retrieving more and Sonnet synthesizing better. We release the full benchmark set\footnote{\href{https://huggingface.co/datasets/nlpai-lab/agenthop}{https://huggingface.co/datasets/nlpai-lab/agenthop}} and the harness\footnote{\href{https://github.com/JohnnyNLP/agenthop}{https://github.com/JohnnyNLP/agenthop}} to support diagnostic agent benchmarking.

\end{abstract}

\input{sections/1.Introduction}

\input{sections/2.Related_work}

\input{sections/3.AgentHop}

\input{sections/4.Experiment}

\input{sections/5.Conclusion}

\bibliographystyle{unsrtnat}
\bibliography{references}


\clearpage
\appendix

\input{sections/Apdx_A} 
\input{sections/Apdx_B} 
\input{sections/Apdx_C}
\input{sections/Apdx_D}
\input{sections/Apdx_E}
\input{sections/Apdx_F}

\input{sections/Apdx_G}


\end{document}

%% file: sections/1.Introduction.tex
\section{Introduction}
\label{sec:introduction}

A \emph{language agent} is a system built on a large language model (LLM) that interacts with its environment by issuing tool calls and reading observations across multiple turns~\citep{sumers2024cognitivearchitectureslanguageagents, yao2023reactsynergizingreasoningacting}. On a multi-step task, such an agent exercises retrieval, reasoning, planning, and execution within a single trajectory~\citep{Wang_2024}, and a failure may originate at any of these stages. Each stage calls for a different fix, and a developer choosing between models for deployment, or directing the next round of post-training, needs to know which stage is binding for which model.

The field has issued repeated calls for behavior-decomposed evaluation~\citep{kapoor2024aiagentsmatter, bean2025measuringmattersconstructvalidity}, yet few benchmarks heed that call. Existing benchmarks for agentic tasks in scientific research~\citep{skarlinski2024languageagentsachievesuperhuman, asai2024openscholarsynthesizingscientificliterature, ren2026scientificintelligencesurveyllmbased}, software engineering~\citep{jimenez2024swebenchlanguagemodelsresolve}, and multi-service workflow~\citep{xu2025theagentcompanybenchmarkingllmagents} have largely inherited the single metric convention from non-agentic evaluation, prioritizing leaderboard comparability over diagnostic insight~\citep{bowman2021will}. Recent decomposition efforts target only one or two axes, such as chain geometry~\citep{you2026agenticragtracerhopawarebenchmarkdiagnosing}, call-level correctness~\citep{he2025trajectbenchatrajectoryawarebenchmarkevaluating}, or per-trajectory progress rate~\citep{ma2024agentboardanalyticalevaluationboard}, leaving cross-axis interactions invisible.

To measure each core agent behavior systematically, we introduce a \textbf{novel four-axis decomposition framework} for agent evaluation that integrates one diagnostic axis per stage into a single per-model report: search for retrieval, synthesis for reasoning, tool-use pattern for planning, and resource management for execution. This exposes each model's binding sub-ability~\citep{bean2025measuringmattersconstructvalidity} and the cross-axis interactions that single-axis benchmarks miss, resonating with \citet{burnell2023rethink}'s broader call for fine-grained reporting in AI evaluation. To put this framework into practice, we built \textbf{AgentHop}, a benchmark tailored to evaluate the four-axis agentic sub-abilities.

Our contributions are: (i) \textbf{AgentHop}, a $1{,}011$-item benchmark for agentic multi-hop question answering, grounded in arXiv citation chains across $7{,}205$ papers and over $450$K citation edges, with seven-tool resource-constrained agentic evaluation; (ii) a \textbf{four-axis decomposition framework} that isolates search, synthesis, tool-use pattern, and resource management from a single evaluation run; and (iii) an \textbf{extensive evaluation of 19 models} that shows per-model strengths and weaknesses, family-level traits, and fine-grained per-axis diagnosis.

%% file: sections/2.Related_work.tex
\input{tables/table1_benchmark_comparison}

\section{Related Work}
\label{sec:related}
We organise prior agent benchmarks around the four ability axes AgentHop measures: agentic search, knowledge synthesis, tool-use pattern, and resource management. We surface key differences below; Table~\ref{tab:benchmark_comparison} summarizes the comparison.

\paragraph{Agentic search.}
Agentic-search benchmarks measure how well a model navigates an open environment to locate evidence. WebArena~\citep{zhou2024webarenarealisticwebenvironment}, GAIA~\citep{mialon2023gaiabenchmarkgeneralai}, and BrowseComp~\citep{wei2025browsecompsimplechallengingbenchmark} score the run end-to-end, without separating navigation skill from synthesis skill at the per-sample level. AgenticRAGTracer~\citep{you2026agenticragtracerhopawarebenchmarkdiagnosing}, the closest concurrent neighbor, decomposes multi-hop failures along \emph{chain geometry}, asking which hop went wrong rather than whether retrieval or reasoning was at fault. AgentHop's independent recall and conversion metrics separate the two, distinguishing a model that retrieves correctly but reasons poorly from one that does the reverse.

\paragraph{Knowledge synthesis.}
Knowledge-synthesis benchmarks measure whether a model can reason across documents already retrieved for it. HotpotQA~\citep{yang2018hotpotqadatasetdiverseexplainable} established multi-hop QA as a canonical two-step retrieval-and-reasoning task; 2WikiMultiHopQA~\citep{ho2020constructingmultihopqadataset} scales hop depth on Wikipedia; OpenScholar~\citep{asai2024openscholarsynthesizingscientificliterature} extends to open-ended scientific synthesis. These benchmarks supply the relevant evidence and evaluate the synthesis step in isolation. AgentHop instead requires the agent to locate the evidence first and then synthesize across it, with the conversion-rate axis isolating reasoning quality from retrieval success.

\paragraph{Tool-use pattern.}
Tool-use benchmarks measure whether a model can invoke external tools correctly: choosing the right function, supplying well-formed arguments, and respecting schema constraints. The Berkeley Function-Calling Leaderboard (BFCL)~\citep{patil2025the} scores single-call and multi-turn function-call correctness, while $\tau$-bench~\citep{yao2024taubenchbenchmarktoolagentuserinteraction} scores end-to-end task completion in multi-turn agent-user-tool interaction; TRAJECT-Bench~\citep{he2025trajectbenchatrajectoryawarebenchmarkevaluating} extends this to trajectory-aware diagnostics on tool selection, argument correctness, and dependency satisfaction. AgentHop adds the per-model action-pattern view via calls-per-turn and per-family bigram and trigram fingerprints, surfacing how each model composes its trajectory rather than just whether each call was well-formed.

\paragraph{Resource management.}
Staying within token, turn, or cost limits is rarely treated as a primary outcome in agent benchmarks. AgentBoard~\citep{ma2024agentboardanalyticalevaluationboard} comes closest by offering progress-rate decomposition alongside accuracy, though it does not budget resources; SWE-bench~\citep{jimenez2024swebenchlanguagemodelsresolve} treats efficiency only secondarily. Beyond these, the deployment-cost dimension is typically reported as supplementary metadata rather than as an evaluation parameter. AgentHop instead treats resource budgets as part of the task's difficulty, decomposing non-submission failures by which cap is binding to attribute how each model spends its budget.

%% file: tables/table1_benchmark_comparison.tex

\begin{table}[t]
\centering
\caption{Comparison of AgentHop with existing benchmarks along the four ability axes and a diagnostic indicator.}
\label{tab:benchmark_comparison}
\small
\setlength{\tabcolsep}{4pt}
\renewcommand{\arraystretch}{1.1}
\newcommand{\cmark}{{\color{green!60!black}\ding{51}}}%
\begin{tabular}{@{}lrccccc@{}}
\toprule
\textbf{Benchmark} & \textbf{N}
  & \textbf{Search}
  & \textbf{Synth.}
  & \textbf{Tool-use}
  & \textbf{Resource}
  & \textbf{Diagnostic} \\
\midrule
HotpotQA \citep{yang2018hotpotqadatasetdiverseexplainable}     & 113K  & \cmark & \cmark & & & \\
2WikiMultiHopQA \citep{ho2020constructingmultihopqadataset}     & 167K  & & \cmark & & & \\
OpenScholar \citep{asai2024openscholarsynthesizingscientificliterature} & 2,967 & \cmark & \cmark & & & \\
WebArena \citep{zhou2024webarenarealisticwebenvironment}        & 812   & \cmark & \cmark & \cmark & & \\
GAIA \citep{mialon2023gaiabenchmarkgeneralai}                   & 466   & \cmark & \cmark & \cmark & & \\
BrowseComp \citep{wei2025browsecompsimplechallengingbenchmark}  & 1,266 & \cmark & \cmark & \cmark & & \\
AgenticRAGTracer \citep{you2026agenticragtracerhopawarebenchmarkdiagnosing} & 1,305 & \cmark & \cmark & \cmark & & \cmark \\
BFCL \citep{patil2025the}                                       & 2,251 & & & \cmark & & \\
$\tau$-bench \citep{yao2024taubenchbenchmarktoolagentuserinteraction} & 165 & & & \cmark & & \\
TRAJECT-Bench \citep{he2025trajectbenchatrajectoryawarebenchmarkevaluating} & 5{,}670 & & & \cmark & & \cmark \\
AgentBoard \citep{ma2024agentboardanalyticalevaluationboard}    & 1,013 & & \cmark & \cmark & \cmark & \\
SWE-bench \citep{jimenez2024swebenchlanguagemodelsresolve}      & 2,294 & & \cmark & \cmark & \cmark & \\
\midrule
\rowcolor{blue!8}
\textbf{AgentHop (Ours)}                                        & 1,011 & \cmark & \cmark & \cmark & \cmark & \cmark \\
\bottomrule
\end{tabular}
\end{table}

%% file: sections/3.AgentHop.tex
\section{The AgentHop Benchmark}
\label{sec:benchmark}

\begin{figure}[t]
\centering
\includegraphics[width=\textwidth]{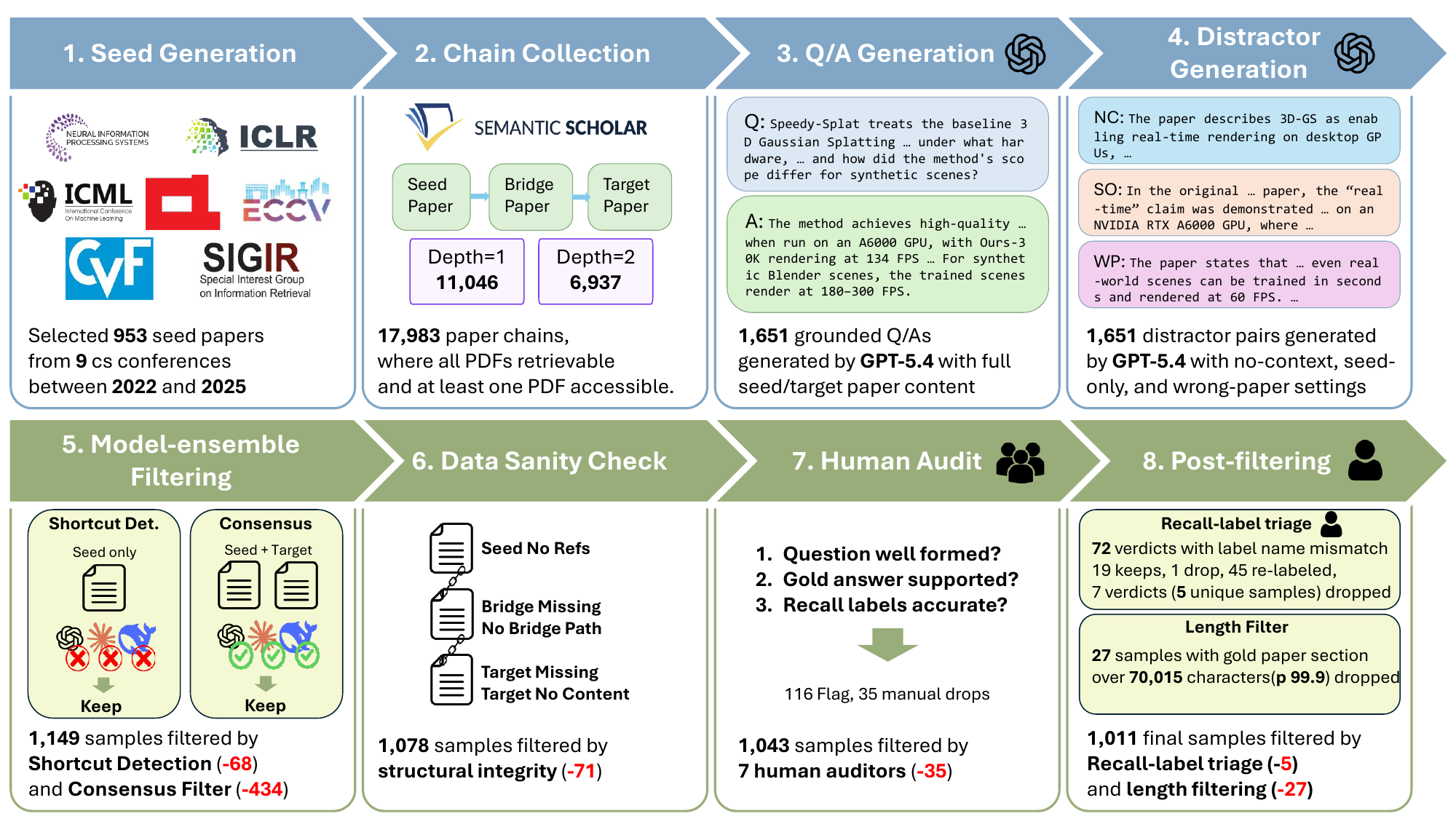}
\caption{Eight-stage construction pipeline. Four \emph{generation} stages (top) seed citation chains from $953$ source papers across nine CS venues, draft grounded question--answer pairs, and assign each pair three engineered distractors. Four \emph{filtering} stages (bottom) retire candidates via an automated 3-model ensemble check, a structural-integrity sanity pass, a 7-auditor human review, and a post-audit recall-and-length pass; $1,011$ items remain.}
\label{fig:pipeline}
\end{figure}

\subsection{Task Definition}
\label{subsec:task}

For each evaluation item, the agent receives a system prompt, the identifier of a single \emph{seed} paper, a research question, and four options labeled A--D. The gold papers that hold the answer are not given; starting from the seed, the agent must navigate the citation graph, read paper sections, and gather evidence before committing to one of the four options. The sandbox of seven built-in tools and the per-run resource budget govern this navigation.

\paragraph{Tools.} The agent operates inside a sandbox of seven tools. Five interact with the literature: \texttt{keyword\_search} retrieves paper IDs whose titles or abstracts match a query; \texttt{get\_paper\_info} surfaces metadata (title, authors, abstract) for a paper ID; \texttt{get\_references} exposes the citation list of a given paper; \texttt{list\_sections} enumerates the section headers of a paper; \texttt{read\_section} returns the full text of a named section. Of these five, only \texttt{read\_section} returns paper content. The remaining two tools control the run: \texttt{think} writes a private deliberation note that does not change the environment, and \texttt{submit\_answer} commits to one of the four options and ends the run. Full function-calling schemas, input arguments, and per-call costs for all seven tools are listed in Table~\ref{tab:tools}.

\paragraph{Resource limits.} Each run is constrained on three axes: a 20-turn cap on model-to-tool exchanges, a 200K-token cap on the aggregated input and output tokens across all turns, and a 30-point tool-call budget from which each tool call subtracts a tool-specific cost. Because \texttt{read\_section} returns the full section text, we set its cost at 5 points to discourage greedy reads. Runs that exhaust any of the three caps before reaching \texttt{submit\_answer} are recorded as \emph{non-submissions}, scored as incorrect, and reported separately as a fail-state diagnostic. When a tool call would overdraw the point budget, the harness rejects the call with an in-context warning and gives the agent one further turn to retry with a cheaper alternative before terminating; full rejection-and-retry semantics are detailed in Appendix~\ref{app:error_handling}. The same protocol applies to every model with no per-model tuning.

\paragraph{Item structure.} Each item is defined by two distinctions. \textbf{Target multiplicity} separates single-target items (ST), with one gold paper, from multi-target items (MT), which require both. \textbf{Depth} separates depth-1 paths, where the agent reaches a gold paper directly from the seed, from depth-2 paths, which pass through one intermediate bridge paper. Crossing the two yields four item types: ST/d1, ST/d2, MT/d1, MT/d2, distinguishing synthesis difficulty from search difficulty. The agent's search space is the two-hop citation neighborhood of the seed, averaging $311$ papers per item; this sets a realistic difficulty regime, large enough that navigation is genuinely challenging yet manageable within the resource budget. 

\subsection{Construction}
\label{subsec:construction}

AgentHop is built from $7,205$ unique research papers connected by over $450$K citation edges, all drawn from nine major computer-science venues over 2022--2025. After an eight-stage pipeline of four automatic-generation stages followed by four filtering and audit stages (Figure~\ref{fig:pipeline}, full detail in Appendix~\ref{app:pipeline}), $1,011$ items survive: $568$ single-target and $443$ multi-target. Each item is a four-option MCQ grounded in specific sections of one or two of these papers. The seed, any bridge, and every gold paper of each surviving item are guaranteed to have parsed section text available to the agent. Other papers in the citation neighborhood may be metadata-only when their arXiv submission lacks HTML-parseable content; \texttt{list\_sections} surfaces this status at 1-point cost before any \texttt{read\_section} attempt. Answering correctly only requires reading gold papers, so partial neighborhood coverage does not block successful trajectories while preserving the full citation graph for navigation. Alongside the correct option, each item carries three engineered distractor options designed to vary along independent dimensions of plausibility: one drawn from the seed paper alone, one drawn from a non-gold paper adjacent to the seed in the citation graph, and one plausible-sounding paraphrase generated by GPT-5.4 without grounding in any supplied paper. We report a detailed analysis of the construction pipeline in Appendix~\ref{app:pipeline}.

\subsection{Evaluation Protocol: Four Diagnostic Axes}
\label{subsec:axes}

Headline accuracy is standard four-option multiple-choice accuracy. In addition, each evaluation run produces four diagnostic measures, defined below, that capture different aspects of the agent's trajectory. For sample $i$, let $r_i \in \{0,1\}$ indicate whether the agent satisfied paper recall, and $c_i \in \{0,1\}$ whether it answered correctly.

\paragraph{Search axis: paper recall.} Each item designates one or more gold papers whose content holds the answer. Paper recall on a sample is $r_i = 1$ iff \texttt{read\_section} returned the content of at least one section of every gold paper of that sample; calls rejected for insufficient budget or malformed arguments do not count, since no content reaches the agent. \texttt{read\_section} is the only tool through which an agent obtains evidence to read.

We additionally report \emph{section recall}, a stricter variant: a sample scores $1$ iff, for every gold paper, a delivered section contains a \texttt{direct}-labelled evidence location. Audit labels cite evidence at the papers' native subsection granularity while the sandbox's readable unit is the top-level section, so labels are resolved to their containing readable section before matching (mapping rule in Appendix~\ref{app:section_recall_mapping}). Table~\ref{tab:main_results} reports both metrics.

\paragraph{Synthesis axis: conversion rate.} Conversion rate is $P(c_i = 1 \mid r_i = 1)$: among samples on which the agent satisfied paper recall, the fraction it answered correctly. Because the denominator is gated on recall, conversion isolates the synthesis ability from retrieval. A model that hits high recall but low conversion can locate the evidence but cannot combine it; the converse profile, low recall and high conversion, identifies a model whose accuracy ceiling is set by retrieval rather than by reasoning.

\paragraph{Tool-use pattern axis: parallel--sequential regime.} The tool-use axis characterizes \emph{how} an agent uses its sandbox via \emph{calls per turn}: the mean number of tool calls emitted per assistant turn. A value of $1.0$ identifies strictly sequential checkpoints that emit one call per turn; any value above $1.0$ implies at least one parallel emission where the model batches related actions within a single turn. The metric has no inherent ``higher is better'' direction; it is read as a regime indicator alongside accuracy. Additionally, we provide tool-call bigram associations and per-family trigram fingerprints in Appendix~\ref{app:trajectory}.

\paragraph{Resource axis: tokens, turns, and budget points.} For each model we report the per-sample means of the three resources as \emph{Avg.\ Tokens}, \emph{Avg.\ Turns}, and \emph{Avg.\ Budgets}, together with three matching \emph{failure rates}: \emph{Max Tokens} marks runs whose aggregate token usage exceeds the 200K cap, \emph{Max Turns} marks runs that reach the 20-turn limit without submitting, and \emph{Max Budgets} marks runs whose tool-call costs exhaust the 30-point budget, each expressed as the percentage of runs that hit the corresponding cap before reaching \texttt{submit\_answer}. Tokens reflect deployment cost, turns reflect decision count, and budget points reflect tool-call accounting; the paired failure rates name which cap binds when a model fails to submit.

%% file: sections/4.Experiment.tex
\section{Experiments}
\label{sec:experiments}

This section evaluates 19 models on AgentHop. Section~\ref{subsec:setup} describes the panel and protocol, Section~\ref{subsec:main_results} answers four diagnostic questions per axis, Section~\ref{subsec:additional} probes three further questions, and Section~\ref{subsec:validity} and Section~\ref{subsec:ablations} close with robustness checks and tool ablations.

\subsection{Setup}
\label{subsec:setup}

We evaluate $19$ models drawn from three capability tiers. The \emph{closed-frontier} tier comprises six commercial checkpoints: Claude Opus 4.6~\citep{anthropic2026claudeopus46} and Sonnet 4.6~\citep{anthropic2026claudesonnet46}, GPT-5.3 codex~\citep{openai2026gpt53codex} and GPT-5.4~\citep{openai2026gpt54}, Gemini-3 Pro~\citep{google2026gemini3pro} and Gemini-3 Flash~\citep{google2026gemini3flash}. The \emph{open-frontier} tier comprises five open-weight checkpoints: DeepSeek-chat and DeepSeek-reasoner~\citep{deepseekai2025deepseekv32pushingfrontieropen}, Moonshot Kimi-K2.5~\citep{kimiteam2026kimik25visualagentic}, MiniMax M2.7~\citep{minimax2026minimaxm27}, and Z.ai GLM-5.1~\citep{zai2026glm51}. The \emph{compute-efficient} tier comprises eight open-weight checkpoints from the Gemma family~\citep{gemmateam2025gemma3technicalreport, gemmateam2026gemma4} (Gemma-3-27B, Gemma-4-31B, Gemma-4-26B-A4B) and the Qwen family~\citep{yang2025qwen3technicalreport, qwen35_27b_dense, qwen35_35b_a3b, qwen36_35b_a3b} (Qwen3-32B, Qwen3-30B-A3B-Instruct, Qwen3.5-27B, Qwen3.5-35B-A3B, Qwen3.6-35B-A3B).

Every model runs the full 1,011 items under the same constraints: seven tools, 20 turns, a 200K-token cap on aggregated tokens across all turns, and a 30-point tool-call budget. Each model is additionally evaluated under a \emph{closed-book protocol} on the same item set: no tools, single-turn answer from prior knowledge, system prompt in Appendix~\ref{app:prompts}, Prompt~9. We use the resulting closed-book accuracy as the zero-tool baseline in the ablation analysis in Section~\ref{subsec:ablations}). 

\subsection{Main Results}
\label{subsec:main_results}

Table~\ref{tab:main_results} shows our main results. We ask four diagnostic questions, one per axis. The answers separate the panel by \emph{family} rather than by tier.

\paragraph{RQ1. Does an agent know when to stop searching?} The answer varies by family. MT recall is consistently lower than ST recall across all 19 models. Looking at what each model does after reaching a gold paper reveals distinct per-family signatures. GPT-5.3 codex and GPT-5.4 \textbf{commit early}: codex submits immediately on $47$\% of ST and $67$\% of MT-both items, and the same policy commits prematurely on $34$--$44$\% of MT failures. Opus 4.6, Sonnet 4.6, and GLM-5.1 \textbf{verify before commit}: nearly all submissions route through \texttt{think}, with Anthropic essentially never raw-submitting ($\leq 2$\%). Gemini-3 Flash, DeepSeek-reasoner, DeepSeek-chat, Kimi-K2.5, and MiniMax-M2.7 \textbf{over-search}: $61$--$89$\% keep reading after the ST gold and $54$--$62$\% after both MT golds; on MT failures, Flash explores without reaching the second gold on $42$\% of items. Gemini-3 Pro stays \textbf{balanced} and is robust on MT failures, with $20$\% premature commit and $38$\% continued exploration. Full per-model decomposition is in Appendix~\ref{app:mt_diag}.

\input{tables/table2_main_results}

\paragraph{RQ2. How do models fail with the correct evidence?} Synthesis abilities vary widely across the panel. Paired same-item tests, as shown in Appendix~\ref{app:paired_tests}, certify the ordering at the extremes: Gemini-3 Pro leads decisively with p=0.0005 against Claude Sonnet 4.6 on common recalled multi-target items and MiniMax-M2.7 trails DeepSeek-reasoner decisively (p=0.0001), while the four checkpoints between them form a cluster that pairwise testing cannot order at $n \approx 200$; we therefore rank conversion only where the tests separate. Crossing recall with conversion partitions the panel into three behavioral profiles. \textbf{Synthesis-bottlenecked} models reach gold papers at open-frontier rates but lose conversion sharply when combining two: MiniMax-M2.7 drops $17$\,pp ST$\to$MT in conversion; Kimi-K2.5 drops $11$\,pp. \textbf{Retrieval-bottlenecked} models show the inverse: Gemma-4-31B converts $0.861$ on the items where it locates both gold papers, near closed-frontier levels, despite a $0.521$ MT recall ceiling. \textbf{Floor models} like Gemma-3-27B and Qwen3-32B fail at paper-reach upstream of either axis, indicating that the legacy training recipe is inadequate for agentic tasks. 

\paragraph{RQ3. Do models act in parallel or in sequence?} Both, with a wide empirical gap between the two regimes. \textbf{Parallel-emitting} checkpoints batch related actions, averaging $1.20$--$1.74$ calls per turn: the Anthropic and OpenAI families, plus MiniMax-M2.7 and GLM-5.1. The remaining checkpoints are \textbf{strictly sequential}, with Gemini-3 Pro and Kimi-K2.5 in the middle at $5$--$7$\% multi-call. Family-specific trigrams sharpen the picture within the parallel regime: Anthropic doubles up \texttt{list\_sections}$\to$\texttt{list\_sections}$\to$\texttt{read\_section} on $\sim 40$\% of trajectories versus a $10.7$\% other-family mean, while OpenAI codex exits-then-relists with \texttt{read\_section}$\to$\texttt{get\_paper\_info}$\to$\texttt{list\_sections} on $30$--$44$\%. Read together, the emission regime and family trigrams act as family fingerprints: a model's tool-call signature places it in a family, layering a provider-level interpretation onto the capability-axis decomposition. These signatures are measured through each provider's native serving interface, so they characterize model-plus-interface systems rather than model architecture alone (Section~\ref{sec:limitations}).

\paragraph{RQ4. How do different models manage the resource constraint?} We find three distinct patterns to AgentHop's tight caps on tokens, turns, and tool-call points. Gemma-3-27B and Qwen3-32B show \textbf{under-engagement}, consuming only $33$--$38$K tokens at accuracy below chance, with $29$\% and $45$\% of their trajectories never calling \texttt{read\_section}; the minimalism reflects give-up, not discipline. Gemini-3 Flash and Gemma-4-26B-A4B \textbf{overspend} the token cap and retire as non-submissions before answering. \textbf{Productive engagement} appears in distinct patterns: GPT-5.4 keeps visible content compact, with no \texttt{think} tool args and minimal assistant text, while Gemini-3 Pro scales resource use with task difficulty; both self-bound well below every cap. Constraint-awareness is what separates the productive cluster from the two pathologies.

\subsection{Additional Analysis}
\label{subsec:additional}

In this section, we address three additional questions to sharpen the per-family attribution: how trajectories distinguish efficient navigators, when models fail to submit, and what within-family upgrades actually improve.

\paragraph{Do tool trajectories speak louder than words?} Yes: trajectory patterns reveal a verification step that separates efficient navigators from the rest~\citep{shinn2023reflexionlanguageagentsverbal}. Bigram associations reproduce the action--observation--think regime of ReAct~\citep{yao2023reactsynergizingreasoningacting}: \texttt{read\_section}$\to$\texttt{think} associates with $+9.8$\,pp mean accuracy, \texttt{think}$\to$\texttt{submit\_answer} with $+10.6$\,pp, and back-to-back \texttt{read\_section} without intervening think with $-6.8$\,pp. Conditional on landing on a non-gold paper, Table~\ref{tab:wasted_read} categorizes what each model does next. \emph{Recover} is rare at $\leq 7$\% across the panel; the discriminative split is between \emph{Think} and the alternatives. Anthropic's Opus 4.6 and Sonnet 4.6 \emph{Think} after $70$--$82$\% of wrong reads and \emph{Stay} only $12$--$13$\%, while GLM-5.1 follows less heavily at $38$\% \emph{Think}. Other closed-frontier checkpoints and DeepSeek-reasoner \emph{Re-nav} on over half of wrong reads; open-weight and compute-efficient checkpoints \emph{Stay} $35$--$45$\% of the time, with Gemma-4-26B-A4B reaching $59$\%; floor models Gemma-3-27B and Qwen3-32B \emph{Submit} on $60$\%+ of their wrong reads, a give-up pattern unique to the floor. Full bigram statistics and the permutation-test detail for the family trigrams are in Appendix~\ref{app:trajectory}.

\input{tables/table3_wasted_read}

\paragraph{When do models fail?} Three distinct mechanisms drive non-submission, each pointing to a different deficit. \textbf{Token-cap exceeded} reflects insufficient self-bounding of trajectory length: sparse-active MoE checkpoints saturate the generation cap within a single turn, led by Gemma-4-26B-A4B at $37$\% and the Qwen MoE family at $7$--$11$\%, and long-trajectory checkpoints accumulate moderate per-turn output across many turns, with Gemini-3 Flash at $28$\% and DeepSeek-chat at $12$\%. \textbf{Max-turns exhausted} reflects an inability to commit: models loop on cheap tool calls without submitting, with Gemma-4-26B-A4B at $7$\%, Qwen3-30B-A3B at $6$\%, Kimi-K2.5 at $4$\%. \textbf{Budget exhausted} reflects unmanaged cost-per-call: parallel-emitters drain the 30-point pool by reading greedily, with GPT-5.4 at $5$\%, MiniMax-M2.7 at $5$\%, GLM-5.1 at $3$\%. Capability tier does not predict any of these: GPT-5.3 codex loses only $1$\% while Gemma-4-26B-A4B loses $44$\%, and Gemma-4-31B in the same compute-efficient tier loses only $2$\%. The decomposition therefore localizes the intervention per failure type: output-length discipline for token-cap failures, commitment training for max-turns, and cost-aware tool-call planning for budget; per-cap decomposition is in Appendix~\ref{app:na_decomp_full}.

\begin{figure}[t]
\centering
\includegraphics[width=\linewidth]{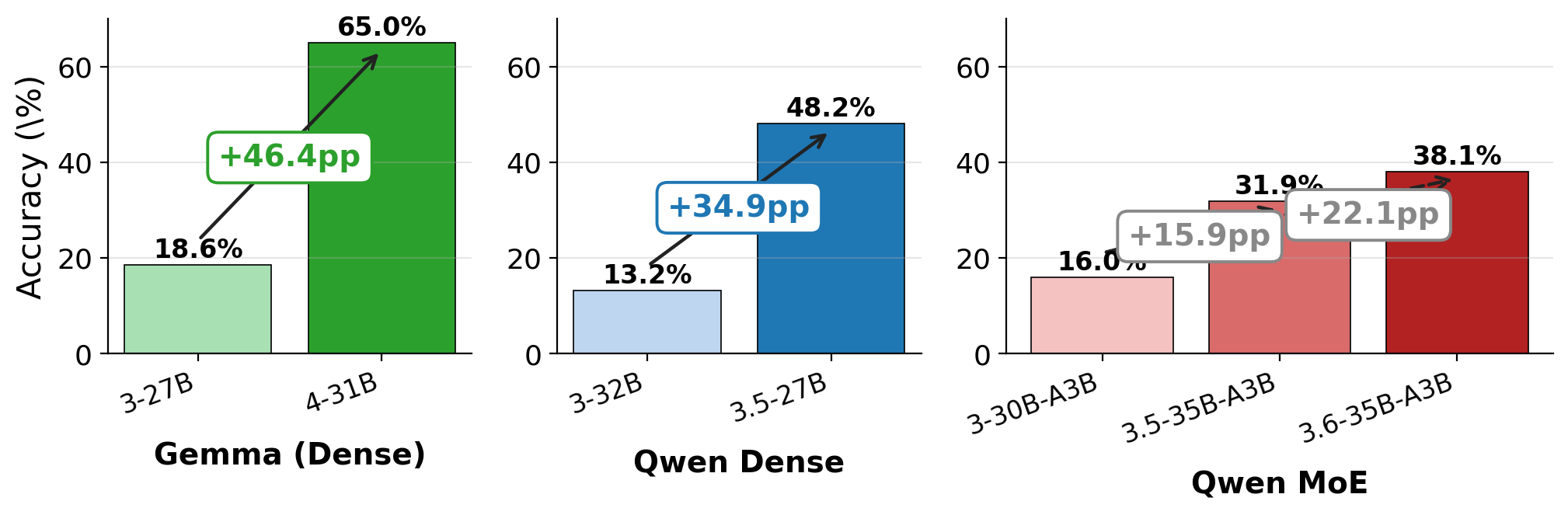}
\caption{Within-family generation gap on overall accuracy. Each panel pairs older-generation checkpoints (light fill) with newer-generation checkpoints (dark fill) within the same architectural cohort, with arrows annotated by the within-family $\Delta$ in percentage points.}
\label{fig:generation_gap}
\end{figure}

\paragraph{Are the newer models better for agentic tasks?} Mostly, but with one caveat. Figure~\ref{fig:generation_gap} pairs older and newer checkpoints within the same family, and the four-axis decomposition exposes tradeoffs that headline accuracy hides. The Gemma~3$\to$4 and Qwen-dense 3$\to$3.5 transitions lift every axis simultaneously, dominated in both cases by a roughly five- to sixfold gain in paper recall: post-training on these lineages is buying \emph{search}. The Qwen MoE lineage 3.5$\to$3.6 tells a different story. Headline accuracy rises modestly, but the lift is driven entirely by synthesis: single- and multi-target conversion both grow, while multi-target paper recall regresses below even the three-generation-old MoE baseline, from $0.472$ to $0.302$. Multi-paper navigation, the one capability the agent loop most directly exercises, is the axis the upgrade quietly gave up.

\subsection{Robustness Checks}
\label{subsec:validity}
Three robustness checks validate AgentHop as a diagnostic instrument: cross-benchmark concordance, internal-consistency reliability, and run-to-run stability.

\paragraph{AgentHop accuracy correlates with reasoning and multi-step benchmarks; less so with domain-specialized ones.} We compute Spearman rank correlations between AgentHop accuracy and seven external benchmarks on the $16$-model overlap. The five reasoning/multi-step benchmarks cluster in a tight $0.67$--$0.73$ band of moderate-to-strong concordance: GPQA-Diamond~\citep{rein2023gpqagraduatelevelgoogleproofqa} ($\rho=0.73$, $n=15$), HLE~\citep{phan2026hle} no-tools ($\rho=0.72$, $n=14$), Terminal-Bench~2.0~\citep{merrill2026terminalbenchbenchmarkingagentshard} ($\rho=0.70$, $n=15$), NL2Repo~\citep{ding2026nl2repobenchlonghorizonrepositorygeneration} ($\rho=0.70$, $n=10$), and HLE with tools ($\rho=0.67$, $n=10$). Two domain-specialized benchmarks correlate more weakly: SWE-Bench Pro/Public~\citep{jimenez2024swebenchlanguagemodelsresolve} ($\rho=0.46$, $n=12$) and BrowseComp~\citep{wei2025browsecompsimplechallengingbenchmark} ($\rho=0.38$, $n=10$). This differential characterizes AgentHop's position in the landscape. The high-correlation group (GPQA, HLE, Terminal-Bench, NL2Repo) places it at the intersection of knowledge, code, and tool-use evaluation, indicating that AgentHop taps the shared agentic-reasoning dimension these benchmarks expose. The lower correlation with domain-specialized benchmarks (SWE-Bench Pro/Public, BrowseComp) confirms AgentHop is not a generic capability proxy: it tracks a distinct competence, locating and combining evidence across cited papers, beyond what the existing landscape covers. Per-benchmark scores are listed in Table~\ref{tab:cross_benchmark} in Appendix~\ref{app:cross_benchmark}.

\paragraph{The instrument exhibits excellent internal-consistency reliability.} Treating the per-sample correctness vector across the $19$ evaluated models as a binary item-response matrix, we compute Cronbach's $\alpha = 0.997$, comfortably above the conventional $\alpha \geq 0.9$ ``excellent reliability'' threshold. Per-item discrimination, measured as the corrected item-total correlation, has a median of $0.55$ (Q1 $0.39$, Q3 $0.67$); $92.5\%$ of items exceed the conventional $r > 0.2$ discrimination threshold and only $3.2\%$ have $r \leq 0$. AgentHop therefore behaves as a well-formed psychometric instrument rather than a noisy aggregate of unrelated questions: most items contribute coherently to the across-model accuracy ranking.

\paragraph{Run-to-run variation is small relative to the cross-model spread, with one exception.} For DeepSeek-chat, Gemma-4-31B, and GPT-5.3 codex we ran the full $1,011$-item evaluation three times. Accuracy standard deviation is $0.009$ for DeepSeek-chat and $0.006$ for GPT-5.3 codex against a between-model spread of $0.758$. Gemma-4-31B is the exception: accuracy std $0.031$ and tokens std $10.2$K (range $21.8$K), the latter comparable to cross-model gaps between adjacent closed-frontier checkpoints. Rank orderings on accuracy and the four axes are preserved across runs, as detailed in Appendix~\ref{app:robustness}.

\subsection{Tool Ablations}
\label{subsec:ablations}

To isolate which tools drive AgentHop's accuracy, we ablate the seven-tool sandbox on three representative checkpoints.

\paragraph{Keyword search contributes only $3$--$10$\,pp on top of reference-following navigation, indicating that the citation graph carries most of the search work.} On three checkpoints (DeepSeek-chat, Gemma-4-31B, GPT-5.3 codex) we run two ablations: \emph{zero-tool} (closed-book) and \emph{no-search} (full sandbox minus \texttt{search\_papers}); protocol in Appendix~\ref{app:ablations}. Table~\ref{tab:ablation_main} shows the decomposition. The no-search variant recovers most of the tool-conditioned lift over the zero-tool baseline ($+23$ to $+35$\,pp); adding \texttt{search\_papers} on top contributes only $+3$ to $+10$\,pp. Reference-following navigation plus content reading delivers the bulk of the accuracy gain; keyword search is dispensable.

\begin{table}[h]
\centering
\caption{Ablation on three checkpoints, with each delta paired between adjacent settings. \emph{Zero-tool}: closed-book. \emph{No-search}: full sandbox minus \texttt{search\_papers}. \emph{Full-tool}: seven-tool sandbox. $\Delta_{\text{tools}}$ is the lift from gaining a tool sandbox (zero-tool $\to$ no-search); $\Delta_{\text{kw}}$ is the additional lift from adding \texttt{search\_papers} (no-search $\to$ full).}
\label{tab:ablation_main}
\small
\setlength{\tabcolsep}{6pt}
\renewcommand{\arraystretch}{1.0}
\begin{tabular}{lcccccc}
\toprule
Model & Zero-tool & $\Delta_{\text{tools}}$ & No-search & $\Delta_{\text{kw}}$ & Full-tool \\
 & (\%) & (pp) & (\%) & (pp) & (\%) \\
\midrule
DeepSeek-chat      & 18.3 & $+35.0$ & 53.3 & $+3.1$  & 56.4 \\
Gemma-4-31B        & 31.7 & $+23.1$ & 54.8 & $+10.0$ & 64.8 \\
GPT-5.3 codex      & 48.6 & $+30.7$ & 79.3 & $+3.2$  & 82.5 \\
\bottomrule
\end{tabular}
\end{table}

%% file: tables/table2_main_results.tex

\definecolor{tierClosed}{HTML}{C0392B}
\definecolor{tierOpen}{HTML}{2471A3}
\definecolor{tierEfficient}{HTML}{27AE60}
\definecolor{headerbg}{HTML}{F0F0F0}
\definecolor{bestcell}{HTML}{E8F5E9}

\definecolor{headerbg}{RGB}{242,242,242}
\definecolor{closedband}{RGB}{230,230,248}
\definecolor{openband}{RGB}{245,240,225}
\definecolor{effband}{RGB}{232,245,232}
\definecolor{bestcell}{RGB}{223,239,223}


\begin{table*}[t]
\centering
\caption{
Main AgentHop results across 19 models, grouped by capability tier and ranked by accuracy within each family. \colorbox{bestcell}{\strut Highlighted} cells mark tier leaders; \textbf{bold} marks the panel best. Axis and failure definitions in Section~\ref{subsec:axes}. Conversion cells are shaded only where a paired same-item test separates the tier leader from the runner-up, as reported in Appendix~\ref{app:paired_tests}.
}
\label{tab:main_results}

\setlength{\tabcolsep}{2.5pt}
\renewcommand{\arraystretch}{0.92}
\setlength{\aboverulesep}{0pt}
\setlength{\belowrulesep}{0pt}
\scriptsize

\resizebox{\textwidth}{!}{%
\begin{tabular}{@{}lccccccccccccccc@{}}
\toprule

\rowcolor{headerbg}
\multicolumn{1}{l|}{\textbf{Model}}
& \multicolumn{4}{c|}{\textbf{Search} $\uparrow$}
& \multicolumn{2}{c|}{\textbf{Synthesis} $\uparrow$}
& \multicolumn{1}{c|}{\textbf{Tool-use}}
& \multicolumn{3}{c|}{\textbf{Resource} $\downarrow$}
& \multicolumn{3}{c|}{\textbf{Failure Ratio} $\downarrow$}
& \textbf{Outcome} $\uparrow$ \\

\rowcolor{headerbg}
\multicolumn{1}{l|}{}
&
\shortstack{\textbf{Recall}\\\textbf{(ST)}}
& \shortstack{\textbf{Recall}\\\textbf{(MT)}}
& \shortstack{\textbf{SecRec}\\\textbf{(ST)}}
& \multicolumn{1}{c|}{\shortstack{\textbf{SecRec}\\\textbf{(MT)}}}
&
\shortstack{\textbf{Conversion}\\\textbf{(ST)}}
& \multicolumn{1}{c|}{\shortstack{\textbf{Conversion}\\\textbf{(MT)}}}
& \multicolumn{1}{c|}{\shortstack{\textbf{Calls}\\\textbf{/Turn}}}
&
\shortstack{\textbf{Avg.}\\\textbf{Tokens}}
&
\shortstack{\textbf{Avg.}\\\textbf{Turns}}
& \multicolumn{1}{c|}{\shortstack{\textbf{Avg.}\\\textbf{Budgets}}}
&
\shortstack{\textbf{Max}\\\textbf{Tokens}}
&
\shortstack{\textbf{Max}\\\textbf{Turns}}
& \multicolumn{1}{c|}{\shortstack{\textbf{Max}\\\textbf{Budgets}}}
&
\textbf{Accuracy} \\

\midrule

\rowcolor{closedband}
\multicolumn{15}{@{}c@{}}{\textbf{Closed Frontier}} \\
\midrule
Gemini-3 Pro            & 0.868 & 0.752 & \cellcolor{bestcell}\textbf{0.859} & \cellcolor{bestcell}\textbf{0.655} & \textbf{0.957} & \cellcolor{bestcell}\textbf{0.970} & 1.05 &  92.4 &  9.4 & 19.8 &  4.4 & 0.1 & 0.0 & \cellcolor{bestcell}\textbf{0.891} \\
Gemini-3 Flash          & 0.724 & 0.501 & 0.680 & 0.386 & 0.625 & 0.644 & 1.01 & 180.3 & 14.1 & 27.9 & 27.7 & 2.1 & 0.2 & 0.547 \\
GPT-5.3 codex           & 0.812 & 0.623 & 0.775 & 0.470 & 0.896 & 0.902 & 1.44 &  69.5 &  8.5 & 25.2 &  0.3 & 0.0 & 0.7 & 0.825 \\
GPT-5.4                 & 0.808 & 0.643 & 0.752 & 0.454 & 0.878 & 0.846 & 1.74 & \cellcolor{bestcell}67.5 & \cellcolor{bestcell}7.7 & 26.5 &  0.3 & 0.0 & 4.9 & 0.790 \\
Claude Opus 4.6         & \cellcolor{bestcell}\textbf{0.873} & \cellcolor{bestcell}\textbf{0.777} & 0.854 & 0.639 & 0.871 & 0.855 & 1.25 & 110.5 & 10.6 & 19.2 &  6.7 & 0.0 & 0.0 & 0.775 \\
Claude Sonnet 4.6       & 0.798 & 0.600 & 0.773 & 0.479 & 0.936 & 0.891 & 1.32 &  95.8 &  9.7 & \cellcolor{bestcell}16.9 &  3.5 & 0.9 & 0.0 & 0.766 \\

\midrule

\rowcolor{openband}
\multicolumn{15}{@{}c@{}}{\textbf{Open Frontier}} \\
\midrule
GLM-5.1                 & 0.768 & \cellcolor{bestcell}0.673 & 0.734 & \cellcolor{bestcell}0.506 & \cellcolor{bestcell}0.830 & 0.768 & 1.20 & 102.0 & 11.5 & \cellcolor{bestcell}23.3 &  3.2 & 1.6 & 3.2 & \cellcolor{bestcell}0.696 \\
DeepSeek-reasoner       & 0.720 & 0.558 & 0.667 & 0.436 & 0.760 & 0.717 & 1.00 & 128.6 & 14.1 & 27.1 &  5.2 & 1.2 & 0.0 & 0.596 \\
DeepSeek-chat           & \cellcolor{bestcell}0.773 & 0.623 & \cellcolor{bestcell}0.743 & \cellcolor{bestcell}0.506 & 0.706 & 0.670 & 1.00 & 156.7 & 15.6 & 26.9 & 12.2 & 2.2 & 0.0 & 0.564 \\
Kimi-K2.5               & 0.764 & 0.569 & 0.720 & 0.463 & 0.613 & 0.504 & 1.08 & 115.5 & 13.3 & 24.7 &  3.6 & 3.9 & 2.5 & 0.441 \\
MiniMax-M2.7            & 0.667 & 0.553 & 0.604 & 0.334 & 0.636 & 0.465 & 1.33 & \cellcolor{bestcell}84.3 & \cellcolor{bestcell}10.5 & 24.6 &  1.0 & 1.4 & 4.8 & 0.413 \\

\midrule

\rowcolor{effband}
\multicolumn{15}{@{}c@{}}{\textbf{Compute-Efficient}} \\
\midrule
Gemma-4-31B             & 0.621 & 0.521 & 0.579 & 0.348 & \cellcolor{bestcell}0.788 & \cellcolor{bestcell}0.861 & 1.01 &  87.6 & 10.9 & 22.6 &  2.4 & 0.1 & 0.0 & \cellcolor{bestcell}0.648 \\
Gemma-4-26B-A4B         & 0.352 & 0.158 & 0.236 & 0.093 & 0.425 & 0.557 & 1.01 & 163.7 & 14.7 & 25.3 & 37.0 & 6.9 & 0.3 & 0.236 \\
Gemma-3-27B             & 0.139 & 0.054 & 0.106 & 0.025 & 0.342 & 0.250 & 1.00 & \cellcolor{bestcell}\textbf{33.6} &  6.5 &  9.5 &  0.1 & 0.0 & 0.0 & 0.186 \\
Qwen3.5-27B             & \cellcolor{bestcell}0.678 & \cellcolor{bestcell}0.589 & \cellcolor{bestcell}0.623 & \cellcolor{bestcell}0.404 & 0.691 & 0.617 & 1.00 & 100.6 & 12.1 & 23.5 &  7.1 & 2.1 & 0.6 & 0.482 \\
Qwen3.6-35B-A3B         & 0.505 & 0.302 & 0.456 & 0.174 & 0.672 & 0.657 & 1.00 & 109.5 & 12.6 & 24.5 & 11.2 & 3.7 & 0.8 & 0.381 \\
Qwen3.5-35B-A3B         & 0.576 & 0.472 & 0.514 & 0.291 & 0.517 & 0.450 & 1.00 & 110.4 & 13.1 & 26.0 &  9.3 & 2.4 & 1.5 & 0.320 \\
Qwen3-30B-A3B           & 0.470 & 0.460 & 0.423 & 0.309 & 0.318 & 0.225 & 1.00 & 101.7 & 12.3 & 18.8 &  4.8 & 6.4 & 0.3 & 0.160 \\
Qwen3-32B               & 0.153 & 0.090 & 0.130 & 0.059 & 0.517 & 0.225 & 1.02 &  37.6 & \cellcolor{bestcell}\textbf{5.3} & \cellcolor{bestcell}\textbf{7.7} &  0.6 & 0.2 & 0.0 & 0.133 \\

\bottomrule
\end{tabular}%
}
\end{table*}

%% file: tables/table3_wasted_read.tex

\begin{table}[t]
\centering
\caption{Action distribution after a wrong-paper \texttt{read\_section} call. Columns: \emph{Recover} reads a gold paper next, \emph{Stay} reads another non-gold section, \emph{Think} invokes \texttt{think}, \emph{Re-nav} calls any other navigation tool, and \emph{Submit} commits an answer immediately.}
\label{tab:wasted_read}

\setlength{\tabcolsep}{3.0pt}
\renewcommand{\arraystretch}{0.92}
\setlength{\aboverulesep}{0pt}
\setlength{\belowrulesep}{0pt}
\scriptsize

\begin{tabular*}{\textwidth}{@{\extracolsep{\fill}}lrrrrrr@{}}
\toprule
\rowcolor{headerbg}
\textbf{Model} & \textbf{n} & \textbf{Recover} & \textbf{Stay} & \textbf{Think} & \textbf{Re-nav} & \textbf{Submit} \\
\midrule

\rowcolor{closedband}
\multicolumn{7}{@{}c@{}}{\textbf{Closed Frontier}} \\
\midrule
Gemini-3 Pro       &   882 & 3.2\%  & 33.9\% & 5.4\%  & 50.2\% & 7.3\% \\
Gemini-3 Flash     &  2165 & 1.2\%  & 39.7\% & 1.5\%  & 50.2\% & 7.0\% \\
GPT-5.3 codex      &  1559 & 1.9\%  & 39.8\% & 0.0\%  & 53.4\% & 4.9\% \\
GPT-5.4            &  1652 & 3.1\%  & 39.6\% & 1.8\%  & 52.1\% & 3.0\% \\
Claude Opus 4.6    &   904 & 1.4\%  & 12.7\% & \textbf{82.3\%} &  3.4\% & 0.1\% \\
Claude Sonnet 4.6  &   723 & 7.3\%  & 11.8\% & \textbf{70.5\%} & 10.1\% & 0.3\% \\

\midrule

\rowcolor{openband}
\multicolumn{7}{@{}c@{}}{\textbf{Open Frontier}} \\
\midrule
GLM-5.1            &  1564 & 1.4\%  & 29.9\% & 38.4\% & 29.0\% & 1.1\% \\
DeepSeek-reasoner  &  1729 & 6.2\%  & 24.3\% &  1.2\% & 58.4\% & 9.9\% \\
DeepSeek-chat      &  1448 & 0.8\%  & 30.3\% & 20.0\% & 42.9\% & 5.9\% \\
Kimi-K2.5          &  1497 & 2.7\%  & 35.5\% & 16.1\% & 37.5\% & 7.4\% \\
MiniMax-M2.7       &  1674 & 6.8\%  & 39.3\% &  9.6\% & 37.3\% & 6.0\% \\

\midrule

\rowcolor{effband}
\multicolumn{7}{@{}c@{}}{\textbf{Compute-Efficient}} \\
\midrule
Gemma-4-31B        &  1749 & 0.7\%  & 38.7\% & 12.6\% & 39.6\% & 8.2\% \\
Gemma-4-26B-A4B    &  2938 & 0.4\%  & 58.8\% & 24.4\% &  8.9\% & 3.3\% \\
Gemma-3-27B        &   405 & 0.7\%  & 20.5\% &  0.2\% & 16.0\% & \textbf{62.2\%} \\
Qwen3.5-27B        &  1441 & 1.9\%  & 37.8\% &  2.8\% & 47.6\% & 8.7\% \\
Qwen3.6-35B-A3B    &  2243 & 1.8\%  & 45.2\% &  3.8\% & 44.1\% & 4.1\% \\
Qwen3.5-35B-A3B    &  1906 & 4.8\%  & 40.8\% &  2.2\% & 41.2\% & 9.9\% \\
Qwen3-30B-A3B      &   678 & 3.1\%  & 44.2\% & 31.9\% & 16.4\% & 3.7\% \\
Qwen3-32B          &   413 & 1.2\%  & 26.6\% &  0.7\% & 10.9\% & \textbf{60.5\%} \\

\bottomrule
\end{tabular*}
\end{table}

%% file: sections/5.Conclusion.tex
\section{Discussion}
\label{sec:discussion}

Across the 19-model panel, agentic behavior on AgentHop separates by model family rather than by capability tier. The four-axis decomposition localizes per-model deficits across search, synthesis, tool use, and resource discipline, surfacing per-family signatures that aggregate accuracy cannot recover.

These diagnoses translate into actionable information for both model developers and pipeline designers. For post-training, the per-model bottleneck attribution names which sub-ability is binding: synthesis-bottlenecked checkpoints benefit from cross-document integration scaffolding, retrieval-bottlenecked checkpoints from navigation training, resource-bound checkpoints from context-management policy. For agentic pipeline design, three regime-level findings distinguish the trade-offs each provider's model-plus-interface system makes: parallel-versus-sequential tool calling, course-correction patterns after wrong reads, and family-clustered failure modes. Together they inform model-selection and trajectory-policy decisions on top of capability tier. Aggregate accuracy compresses these distinctions into one number; per-axis attribution is the natural next step as agent evaluation matures from leaderboard verdicts to mechanism-level diagnosis.

%% file: sections/Apdx_A.tex
\section{Limitations}
\label{sec:limitations}

Despite a rigorous data pipeline and evaluation process, AgentHop is not without limitations. Five caveats qualify the conclusions of the main paper: a parametric-knowledge floor, a single-domain corpus, generator-evaluator overlap, differential axis informativeness, and the observational nature of the trajectory findings.

\paragraph{Parametric floor.}
AgentHop evaluates models that have largely been trained on the cited literature; some fraction of each model's accuracy is parametric familiarity rather than agentic competence. The Section~\ref{subsec:ablations} zero-tool ablation provides one bound: closed-book accuracy ranges $0.49$--$0.73$ for closed-frontier and $0.18$--$0.38$ for open-frontier checkpoints, while the no-search lift over closed-book is $+23$ to $+35$\,pp on the three ablated models. We do not subtract this floor counterfactually; the four-axis decomposition surfaces per-model behavior on top of the floor rather than offering a counterfactual measure of agentic capability.

\paragraph{Single domain, single format.}
The corpus is restricted to nine computer-science venues over 2022--2025, and every item is a four-option multiple-choice question. Behaviour on other fields, on long-tail venues, and on open-ended response formats is not measured here, and we do not claim that the per-model bottleneck attributions generalize outside this distribution.

\paragraph{Generator--evaluator overlap.}
GPT-5.4 is used to generate the questions, the distractor options, and the audit-prep guides for all 1,011 released items. To bound the effect of generator identity, we rebuilt the pipeline with a disjoint generator (Inkling) on fresh seeds and obtained 94 items, of which 91 remain after the same human audit and triage; they are released as a separate split. GPT-5.4 scores 79.1\% on the subset it did not write, with a 95\% confidence interval of 69.7--86.2, and 79.0\% on the items it generated, a difference of 0.1 percentage points. However, the subset is small and therefore we cannot exclude subtler stylistic overlap and does not support per-axis comparison.

\paragraph{Differential axis informativeness.}
The four axes are not equally informative across the capability spectrum. Paper recall is most discriminating in the middle and bottom of the panel, where the difference between $0.10$ and $0.83$ recall is enormous; at the top, the closed-frontier paper-recall range ($0.71$--$0.83$) is narrow. Conversion rate, by contrast, separates closed-frontier checkpoints sharply (Pro $0.96$, Opus $0.86$) but compresses at the bottom where recall denominators are small. We do not claim equal diagnostic power on every axis at every tier; we expect different axes to carry the load for different models.

\paragraph{Inference confounds.}
Tool-call signatures such as calls per turn, think usage, and family trigrams are measured through each provider's native serving interface. Reasoning modes, parallel tool-call support, and tool-call parsing differ across providers and are not controlled. The family fingerprints in Section~\ref{sec:experiments} therefore characterize model-plus-interface systems, not model architecture alone.

\paragraph{Observational, not causal.}
The behavioral findings are correlational. Trigram-pattern signatures associate with model families and strategy concentration associates with capability tier, but trajectory data alone cannot establish that a particular tool-use pattern \emph{causes} a particular accuracy level. We treat strategy--accuracy associations as diagnostic correlates rather than as architectural prescriptions.

%% file: sections/Apdx_B.tex

\section{Ethical Considerations}
\label{app:ethics}

We discuss ethical considerations and broader impacts of AgentHop as follows.

\paragraph{(1) Intellectual Property.}
AgentHop is built from $7{,}205$ public arxiv papers across nine major computer-science venues (2022--2025), each under its respective arXiv-deposited license (predominantly arXiv-perpetual and CC-BY family). We use only paper identifiers, abstracts, and named-section text, consistent with research-fair-use of publicly available scholarly content; we do not reproduce figures, tables, or any non-textual content. Closed-frontier checkpoints (Anthropic Claude, OpenAI GPT, Google Gemini) are accessed under each provider's API terms of service. Open-frontier checkpoints are accessed via two API providers: \texttt{deepseek-chat} and \texttt{deepseek-reasoner} (both resolving to DeepSeek-V3.2~\citep{deepseekai2025deepseekv32pushingfrontieropen} during our evaluation window) via DeepSeek's own API; Kimi-K2.5, MiniMax-M2.7, and GLM-5.1 via the OpenAI-compatible Together.ai endpoint. Compute-efficient open-weight checkpoints (Qwen3.x, Gemma~3/4) are served locally on a pair of H100 GPUs with vLLM. All evaluations use the model checkpoints solely for academic comparison and respect each provider's terms of use. AgentHop will be released on Hugging Face under the CC-BY~4.0 license; the evaluation harness will be released on GitHub under the Apache-2.0 license, supporting reproduction of reported results by running models against the same protocol under the same resource budget.

\paragraph{(2) Broader Impacts.}
AgentHop is a diagnostic instrument for evaluating agents on multi-hop question answering over scientific citation graphs. AgentHop scores can be reported as a leaderboard ranking; for deployment decisions, we recommend pairing them with established benchmarks for the target task and consulting the per-axis decomposition of Section~\ref{subsec:axes} for failure-mode attribution. To preserve evaluation integrity, the benchmark items and gold-section labels should not be incorporated into LLM training corpora.

\paragraph{(3) Controlling Potential Risks.}
AgentHop items are multiple-choice questions over public scholarly content. The released dataset contains no personal information, no harmful or offensive content, and no generative model weights. The evaluation harness operates as a closed sandbox: tools are limited to the seven literature-interaction primitives in Section~\ref{subsec:task} and do not perform external API calls or web access beyond model inference itself.

\paragraph{(4) Human Annotation.}
Seven auditors (two master's-level and five PhD-level computer-science students) reviewed audit-prep items during Stage~7 of the construction pipeline described in Appendix~\ref{app:pipeline}. Auditors are members of the research team and participated voluntarily as part of the paper's data-curation process; no external paid annotation was conducted. Each auditor used the audit-prep guide reproduced in Appendix~\ref{app:prompts} and the Gradio review interface to inspect machine-generated items, the relevant source-paper sections, and the proposed gold answer. No personal information of auditors is collected or released.

\paragraph{(5) LLM Usage.}
LLMs are used in three distinct roles in this work. \emph{Construction:} GPT-5.4 is the primary generation model for question--answer pairs (Stage~3) and audit-prep items (Stage~7); the three-model shortcut-detection and consensus-filter ensemble in Stage~5 uses GPT-4.1, Claude Sonnet~4.6, and \texttt{deepseek-chat} (full prompts in Appendix~\ref{app:prompts}). \emph{Evaluation:} the 19-model panel of Section~\ref{sec:experiments} is the subject of the evaluation and is not used to generate or label test data after Stage~7; the generator--evaluator overlap involving GPT-5.4 is acknowledged as a limitation in Section~\ref{sec:limitations}. \emph{Writing assistance:} LLMs were used for editing and formatting only and did not contribute to scientific claims.

%% file: sections/Apdx_C.tex

\section{Construction Pipeline Details}
\label{app:pipeline}

This appendix expands the eight-stage construction pipeline summarized in Section~\ref{subsec:construction}. Table~\ref{tab:funnel} records the per-stage drop counts. The pipeline is symmetric by design: four \emph{generation} stages assemble each candidate sample (seed $\to$ citation chain $\to$ Q/A $\to$ distractors) and four \emph{filtering} stages then retire candidates that fail one or another quality criterion (model-ensemble checks $\to$ structural and content sanity $\to$ human audit $\to$ post-audit triage).

\begin{table}[h]
\centering
\caption{Per-stage attrition through the eight-stage construction pipeline. Stages 1--4 grow each sample from a seed paper to a fully populated four-option MCQ; Stages 5--8 then filter the resulting pool.}
\label{tab:funnel}
\small
\setlength{\tabcolsep}{8pt}
\begin{tabular}{r l r r}
\toprule
\textbf{Stage} & \textbf{Description} & \textbf{Drops} & \textbf{Cumulative} \\
\midrule
\multicolumn{4}{l}{\emph{Generation}} \\
1 & Seed selection                              & ---     &     953 \\
2 & Citation-chain expansion                    & ---     & 17{,}983 \\
3 & Q/A generation (BFS + DFS)                  & ---     &  1{,}651 \\
4 & Distractor generation (3-trap)              & ---     &  1{,}651 \\
\midrule
\multicolumn{4}{l}{\emph{Filtering}} \\
5 & Model-ensemble filtering (shortcut detection $-68$ + consensus $-434$) & $-502$ & 1{,}149 \\
6 & Data sanity check (structural integrity)    & $-71$   &  1{,}078 \\
7 & Human audit (116 flagged, 35 manually retired) & $-35$ & 1{,}043 \\
8 & Post-filtering (recall triage $-5$ + length filter $-27$) & $-32$ & \textbf{1{,}011} \\
\bottomrule
\end{tabular}
\end{table}

\paragraph{Stage 1: Seed selection.}
Venue and citation metadata are obtained from the Semantic Scholar API. The 953 seed papers are stratified across nine venues, NeurIPS, ICML, ICLR, ACL, EMNLP, NAACL, CVPR, ECCV, SIGIR, spanning publication years 2022--2025 to represent contemporary methodological breadth. Within each venue, seeds are sampled in three year-tiers with citation-count thresholds calibrated so that older papers must clear a higher bar to be included: \emph{recent} (2024--2025, no citation requirement, target 500), \emph{established} (2023, $\geq 50$ citations, target 300), and \emph{well-known} (2022, $\geq 300$ citations, target 200). The decreasing-recency, increasing-citation rule keeps the seed pool dominated by genuinely recent work while ensuring that the older tier still admits only papers that have demonstrably sustained community attention. Stratification balances per-venue counts within $\pm 15\%$ of the target proportion, requires each seed to have 15--50 references (enough for chain construction, while excluding surveys), and removes position and tutorial papers via metadata filtering.

\paragraph{Stage 2: Citation-chain expansion.}
For each seed we follow outgoing citations one or two hops, retaining only chains in which every paper is recoverable through Semantic Scholar API and has at least one accessible PDF. Each candidate chain is then scored by a hierarchical filter: a Jaccard reference-overlap pre-filter (threshold $\geq 0.08$ on shared references between adjacent hops, ensuring topical coherence), a citation-context substantiveness check ($\geq 80$-character methodology- or result-intent contexts; ``background''-only citations are skipped), and finally an LLM judge that rates the \emph{curiosity gap} of the chain on a 1--5 scale conditioned on abstracts and citation context. The judge prompt (Appendix~\ref{app:prompts}, Prompt~\ref{prompt:chain_judge}) instructs GPT-5.4 to reason about whether a researcher reading the seed would naturally want to follow the citation to read the candidate paper, and only chains scoring $\geq 4$ survive. Chains with broken edges, single-author dominance, or cited-paper publication years before 2020 are dropped. The 953 seeds yield 17{,}983 candidate chains: 11{,}046 at depth-1 and 6{,}937 at depth-2.

\paragraph{Corpus collection.}
After candidate chains are constructed in Stage~2, full text for every paper in the surviving citation neighborhoods is fetched and parsed. The arXiv API supplies titles, authors, abstracts, and references; body content is obtained from ar5iv (the arXiv HTML rendering service) for section-structured parsing, and falls back to arXiv-hosted HTML when ar5iv rendering fails. The fetched HTML is parsed into named sections and cached. Papers without an accessible arXiv source, conference accepted papers whose authors did not deposit a preprint, or older venues with gated proceedings, are retained as metadata-only nodes in the citation graph: \texttt{get\_paper\_info} and \texttt{get\_references} continue to operate on them, while \texttt{read\_section} returns a not-available signal. The nine venues used in Stage~1 are chosen for high arXiv-deposit rates, ensuring stable section-level coverage of the gold path. Across the $1{,}011$ retained samples, two-hop neighborhoods average $311$ papers per query (median $315$, range $82$--$667$), of which a median $137$ carry parsed section text (range $8$--$274$). The seed-to-bridge-to-target chain of every sample is guaranteed by the Stage~6 structural integrity check to have full content; metadata-only papers elsewhere in the neighborhood reflect realistic arxiv coverage gaps, and dead-end \texttt{read\_section} calls on them surface in the tool-use and resource axes. Empirically, neighborhood size does not predict difficulty: across the $1{,}011$ retained samples, the Spearman correlation between two-hop neighborhood size and panel-19 mean correctness is $+0.01$ ($p=0.70$, n.s.), with per-model correlations all within $|\rho|<0.07$. Pool size is therefore neighborhood breadth, not difficulty.

\paragraph{Stage 3: Q/A generation.}
GPT-5.4 generates one open-ended question--answer pair per chain that survives the pre-filters of Stage~2, following templated prompts that target the citation evidence (single-target / BFS) or the synthesis between two cited papers (multi-target / DFS); the verbatim system prompts are reproduced in Appendix~\ref{app:prompts} (Prompt~\ref{prompt:bfs_gen} for BFS and Prompt~\ref{prompt:dfs_gen} for DFS). The prompts explicitly forbid the generator from naming the target paper's method names or title verbatim in the question, so an agent cannot shortcut to the gold paper by pattern-matching the question against paper titles or methodology vocabulary. Each call receives the full text of the seed paper, the gold target paper(s), and the chain edge metadata, and emits a structured record containing the question, the gold answer, and the verbatim answer-context span from the cited paper. The generator additionally tags each item with one of four \emph{reasoning-type} labels reflecting what the question asks for: \textbf{GROUND} (a paper-grounded factual claim), \textbf{METHOD} (a specific technical procedure used by a paper), \textbf{MOTIVE} (why a paper was written or how it relates to prior work), and \textbf{RESULT} (an empirical outcome the paper reports). The labels are assigned at generation time from the question structure and verified during Stage~7; they are not used to score answers and do not enter the four-axis decomposition of Section~\ref{subsec:axes}, but supply a categorical breakdown reported in Appendix~\ref{app:reasoning_type}. Generations that violate the schema or fail the seed-grounding rules are retried up to three times before the chain is dropped. Stage~3 yields 1{,}651 grounded Q/A pairs ready for distractor generation, with reasoning-type counts of GROUND $330$, RESULT $292$, MOTIVE $265$, and METHOD $124$ in the final $1{,}011$-item set.

\paragraph{Stage 4: Distractor generation.}
Each Q/A pair is then expanded into a four-option MCQ by generating the three diagnostic distractors of Section~\ref{subsec:construction}, \textbf{seed\_only} and \textbf{wrong\_paper}, both produced from Prompt~\ref{prompt:mcq_grounded}, a paper-grounded prompt where the generator answers from the seed paper or a sibling reference, and \textbf{no\_context}, produced from Prompt~\ref{prompt:mcq_parametric}, a parametric prompt where the generator answers from internal knowledge alone. The generator never sees the words ``distractor'' or ``wrong answer''; it genuinely tries to answer each question from the wrong or absent source, and the answer is incorrect because the source is wrong. Distractors with high lexical overlap against the gold answer are rejected so that the four options remain semantically separable. Stage~4 retains all 1{,}651 candidates, each now carrying a gold answer plus three diagnostic distractors.

\paragraph{Stage 5: Model-ensemble filtering.}
The 3-model ensemble (GPT-4.1, Claude Sonnet 4.6, \texttt{deepseek-chat}) runs two filters with complementary intents.

The \emph{shortcut detection} pass, run with Prompt~\ref{prompt:filter_ab}, checks whether \emph{additional retrieval is actually required}. Each model is given the seed paper text alongside the question and four options; if the ensemble can recover the gold answer from the seed alone, the sample is leaking its answer into the seed and the agent could shortcut past the cited target paper at evaluation time, so the sample is dropped (68 samples).

The \emph{consensus filter}, run with Prompt~\ref{prompt:filter_c}, checks whether the gold answer is \emph{the most evident option once the cited paper is provided}. Each model is given the seed paper plus the gold target paper(s) alongside the question and options. We drop samples on which the ensemble cannot converge on the gold answer with full paper access (434 samples), these are weak-gold cases where, even with the cited paper in front of the model, a distractor is more parametrically plausible than the gold itself, so the gold would not stand out as the correct choice at evaluation time. This filter also retires samples whose seed-only or no-context distractor turns out to be factually correct on its own merits, since the ensemble cannot converge on the labeled gold in such cases.

\begin{figure}[t]
\centering
\includegraphics[width=\linewidth]{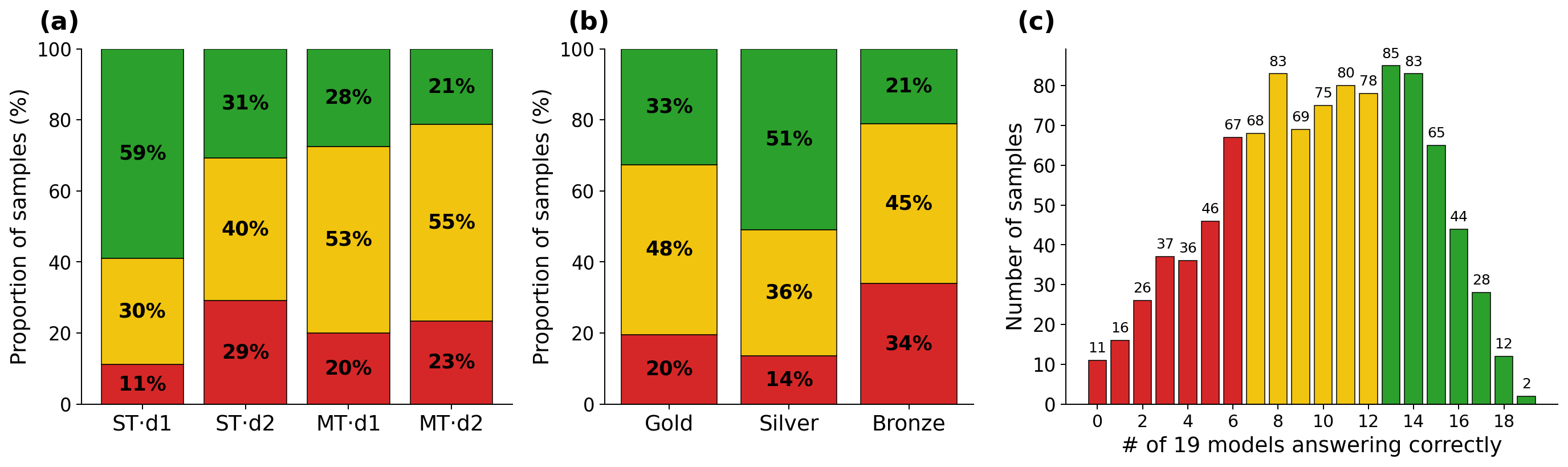}
\caption{Sample-difficulty distribution across the $1{,}011$-sample benchmark. Each sample is bucketed by how many of the $19$ evaluated models answered it correctly: low ($0$--$6$, red), mid ($7$--$12$, yellow), high ($13$--$19$, green). \textbf{(a)} Stratified by question type $\times$ depth. \textbf{(b)} Stratified by consensus tier. \textbf{(c)} Per-sample coverage histogram, with bar color matching the low/mid/high coding.}
\label{fig:difficulty_distribution}
\end{figure}

The 502 dropped samples are recorded with their disagreement patterns. The retained 1{,}149 samples are also assigned a consensus tier from these patterns: \textbf{Gold} requires shortcut detection 0/3 and consensus 3/3 (no model could shortcut to the gold from the seed alone, and all three converge on the gold once the target paper is provided), \textbf{Silver} requires shortcut detection 1/3 with consensus 3/3, and \textbf{Bronze} covers consensus 2/3 (one of the three models disagrees on the gold). Empirically the tiers index construction rigour rather than designed difficulty: mean correctness across the six closed-frontier models is 81.8\% on Silver, 79.3\% on Gold, and 70.5\% on Bronze. Silver outperforms Gold because Silver samples remain partially seed-answerable and so are easier in absolute terms, while Bronze is the hardest because the consensus filter isolates closer-call cases. The principled difficulty axes are question type and depth (Figure~\ref{fig:difficulty_distribution}, panel~(a)).

\paragraph{Stage 6: Data sanity check.}
Stage~6 enforces \emph{structural integrity} of every retained sample's gold path and retires 71 candidates. Six structural defects, any one of which retires the sample, are checked: \texttt{SEED\_NO\_REFS} (the seed has no outgoing references in the citation graph); \texttt{BRIDGE\_MISSING} (a depth-2 BFS bridge node does not exist in the graph); \texttt{NO\_BRIDGE\_PATH} (no seed reference leads to the terminal paper); \texttt{TERMINAL\_GHOST} (the terminal paper exists in the graph but lacks title metadata); \texttt{TERMINAL\_NO\_CONTENT} (the terminal has no body sections in the released paper pool); and \texttt{TARGET\_GHOST} / \texttt{TARGET\_NO\_CONTENT} (the multi-target analogues for DFS). All 71 drops at this stage fall into one of these six codes, leaving 1{,}078 audit-ready candidates.

Two further automated checks run alongside the structural integrity check but do \emph{not} drop samples directly. A \emph{content-navigability} check verifies that distinctive keywords (numbers, proper nouns, technical terms) extracted from the gold answer appear in the gold-target paper's body text; samples that fail are recorded as warnings for the Stage~7 auditors but are not retired automatically. A separate \emph{appendix-grounding} check flags samples whose gold answer fragments are found only in appendix sections rather than the main body. We deliberately do not retire appendix-grounded samples here, some of our gold answers genuinely live in supplementary material, and instead surface the flag as a hint to Stage~7 auditors who decide whether the appendix grounding is acceptable on a case-by-case basis.

\begin{figure}[h]
\centering
\includegraphics[width=\textwidth]{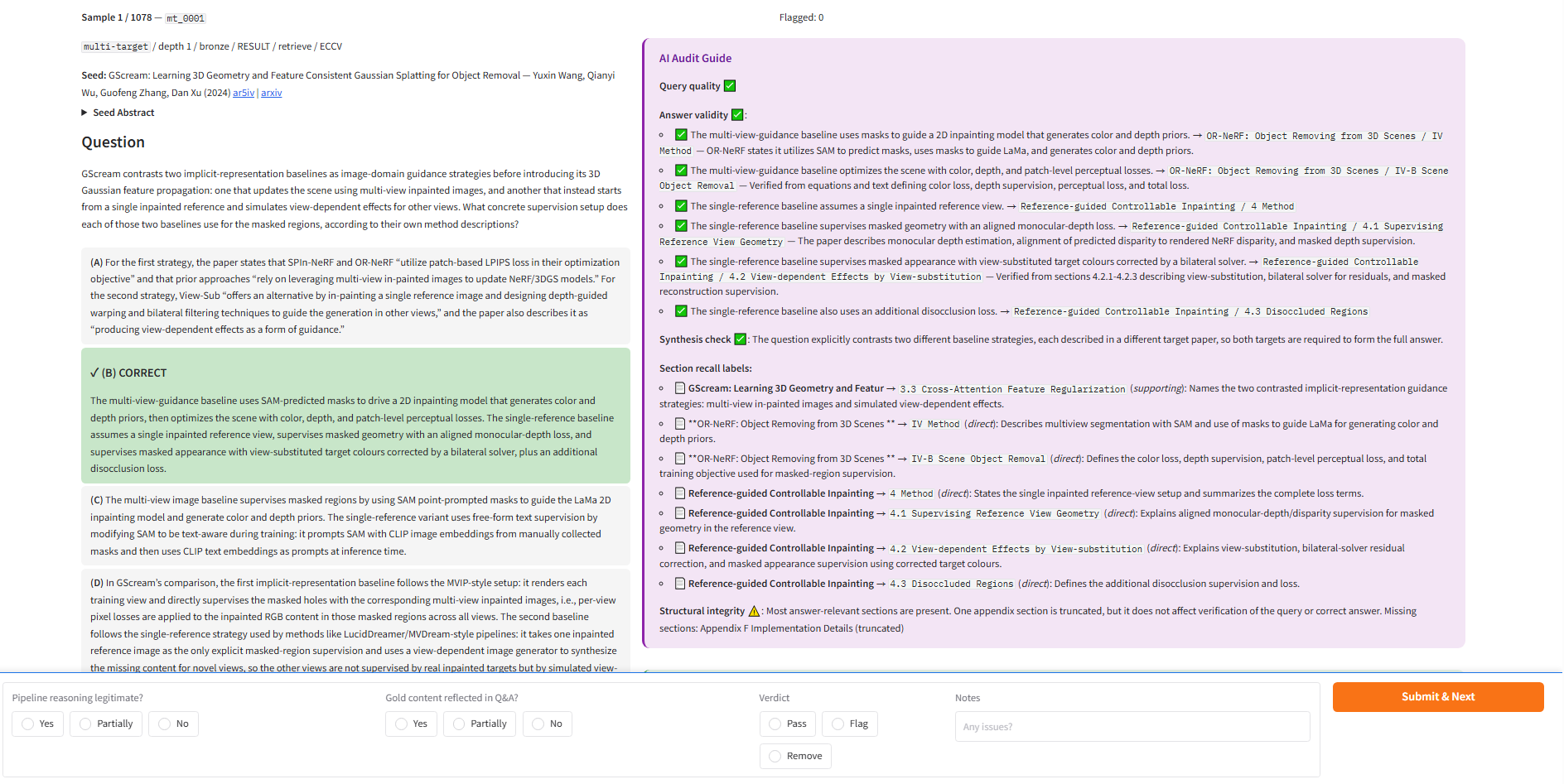}
\caption{Gradio human-audit interface used in Stage 7. Each sample is presented with the question, four options (gold and three distractors), the GPT-5.4 audit JSON, and links to underlying papers. Auditors mark three quality axes and submit one of three verdicts. The same interface is released alongside the benchmark, constrained to the seven-tool, 30-point budget environment, for future human-vs-LLM head-to-head studies.}
\label{fig:audit_interface}
\end{figure}

\paragraph{Stage 7: Human audit.}
The 1{,}078 samples that pass Stage 6 are reviewed by seven auditors through the Gradio interface shown in Figure~\ref{fig:audit_interface}. Each sample is presented with the question, the four options, the GPT-5.4 audit-prep JSON described below (under \emph{Audit-prep generation}), and links to the underlying paper PDFs. Auditors mark three quality axes: (i) whether the question is well-formed and unambiguously answerable from the cited papers, (ii) whether the gold answer is supported by the cited evidence, and (iii) whether the section-recall labels are accurate (with the option to add, remove, or re-classify labels). They submit one of three verdicts, \texttt{Pass}, \texttt{Flag} (with a free-text note), or \texttt{Remove}; the latter two collapse to a single \emph{flagged} category at this stage and are not retired automatically. Across the 1{,}078 samples, 116 receive at least one non-Pass verdict and the remaining 962 are kept as Pass.

The 116 flagged samples are then reviewed in a follow-up triage pass by the lead author, who decides per-sample whether to retain or retire each one. 35 of the 116 are retired (the remaining 81 are retained with the flag preserved as metadata). Table~\ref{tab:audit_drop_themes} decomposes the 35 retired samples by auditor-note theme: roughly half are infrastructure issues that the automated Stage~6 sanity check missed (HTML-parser collapses or section-extraction errors on the gold paper), and the rest split between weak-chain / under-utilised target, trivial or low-quality question, and answer-not-addressing-question alignment problems.

\begin{table}[h]
\centering
\caption{Themes among the 35 samples retired at Stage~7 after manual triage of the 116 audit-flagged candidates. The largest cluster is paper-content extraction failures that the automated structural-integrity check at Stage~6 could not catch because they occur deep inside section text rather than at the metadata level.}
\label{tab:audit_drop_themes}
\small
\setlength{\tabcolsep}{8pt}
\begin{tabular}{l r p{0.55\textwidth}}
\toprule
\textbf{Theme} & \textbf{$n$} & \textbf{Description} \\
\midrule
Paper-content extraction broken               & 12 & The gold paper's HTML or section-extraction pipeline produced empty, mangled, or partially captured sections; the auditor saw that the cited paper content needed to ground the answer was missing. \\
Weak chain / target paper under-utilised      &  6 & The synthesis relies on only one of the two cited papers, or the bridge $\to$ terminal step does not actually carry the information the question demands. \\
Trivial or low-quality question               &  6 & The question or its gold answer is too simple to be diagnostic of multi-hop competence (e.g., the answer is a single-word fact derivable without genuine multi-paper reasoning). \\
Answer-question alignment failure             &  6 & The gold answer is on-topic but does not satisfy what the question explicitly asks (\emph{how} vs \emph{what}, missing tradeoff or comparison, synthesis claim that doesn't follow from the cited evidence). \\
\midrule
Other (sundry, 1 each)                        &  5 & Option text misquotes the paper; a numerical detail is incorrect; a non-gold option is also defensible; the answer compares the wrong pair of methods; or the asserted target-paper claim actually originates in the seed paper. \\
\midrule
\textbf{Total retired at Stage 7} & \textbf{35} & \\
\bottomrule
\end{tabular}
\end{table}

\paragraph{Audit quality report.}
The audit team comprised seven volunteer auditors recruited from the authors' research group, anonymised as Auditor A through G. All seven are graduate students in computer science (5 PhD candidates, 2 MS students) at the authors' institution, with prior research experience in machine learning or NLP. Participation was voluntary and uncompensated, and the audit served as internal validation of a benchmark released alongside this paper.

Before the audit began, all auditors received a 30-minute in-person walk-through of the evaluation logics, the three quality axes, and several worked examples of \texttt{Pass}, \texttt{Flag}, and \texttt{Remove} cases. Auditors were instructed to spend at most three minutes per sample to keep throughput steady and avoid fatigue; if a sample required longer the auditor was asked to flag it for review rather than continue.

Across the 1{,}078 samples we recorded 1{,}204 review events, with the additional events corresponding to auditor-initiated revisits (e.g., reopening a flagged sample after consulting the paper). Per-sample wall-clock time, measured between the moment a sample is presented and the moment a verdict is submitted, has a long tail. We treat any single review longer than 300 seconds as away-from-keyboard (AFK) and exclude 187 such events from the timing summary. On the remaining 1{,}017 engaged-time events, the median per-sample audit time is \textbf{88 seconds} (mean 97 seconds, p95 228 seconds), implying a per-auditor throughput of roughly 30--40 reviewed samples per hour at full engagement. Per-auditor medians range 56--211 seconds; we observe no systematic relationship between speed and verdict severity.

We prioritized full single-pass coverage, every one of the 1{,}078 samples reviewed by exactly one auditor, over partial coverage with designed double-coding for inter-rater agreement statistics, given the seven-auditor capacity budget. Consistency across auditors is therefore enforced procedurally rather than statistically: through the structured Gradio interface shown in Figure~\ref{fig:audit_interface}, the GPT-5.4 audit-prep guide (next paragraph) which front-loads claim decomposition and section-recall hypotheses, and a fixed three-axis verdict schema (\texttt{Pass}/\texttt{Flag}/\texttt{Remove}) with required free-text notes on \texttt{Flag}. We acknowledge this trade-off as a limitation of the present audit design.

\paragraph{Audit-prep generation.}
Each post-Stage-6 sample is paired with a structured audit JSON produced by GPT-5.4, which serves both as a verification record and as a guide that orients human reviewers. The same call decomposes the gold answer into atomic factual claims and verifies each against the cited sources (\emph{answer\_validity}); checks that the seed-to-bridge-to-target chain is necessary and well-formed for single-target samples (\emph{chain\_coherence}); checks that the answer genuinely synthesizes information from both targets for multi-target samples (\emph{synthesis\_check}); and emits a list of \texttt{section\_recall\_labels} naming the section headers that contain evidence for the gold answer. The structured artefact speeds the audit and enforces a uniform claim decomposition across auditors; in practice it expedited per-sample check time substantially and helped auditors stay focused throughout the session. The full audit prompt is Prompt~\ref{prompt:audit} in Appendix~\ref{app:prompts}.

\paragraph{Section-recall labels.}
Each \texttt{section\_recall\_label} pairs a section header from the gold target paper(s) with a relevance class: \texttt{direct} (the claim is stated explicitly in this section) or \texttt{supporting} (the section provides context needed to interpret the claim). Multiple sections may carry either class for a single sample, and a sample may have $\geq 1$ section per gold paper. The headline search-axis metric in the main text is \emph{paper recall}, the agent reads at least one section of every gold paper. \emph{Section recall}, the stricter variant requiring a delivered section to contain a  \texttt{direct}-labelled evidence location, is reported alongside paper recall in Table~\ref{tab:main_results}. Subsection labels are resolved to their containing readable section as described in Appendix~\ref{app:section_recall_mapping}. Auditors may add, remove, or re-classify labels during Stage 7; the released labels are post-audit and post-Stage-8.

\paragraph{Stage 8: Post-filtering.}
A final two-step pass after the human audit drops 32 more samples.

\textbf{Step 1: Recall-label triage (manual, $-5$ samples).} The Stage~7 audit produces a section-recall label for every retained sample (the gold paper section in which each atomic claim from the gold answer is verifiable). The audit-prep guide that drafts these labels can mis-name a section header, point at a child header that does not exist verbatim in the released paper pool, or flag a claim as unverified at the labeled section even though the claim is in fact present elsewhere in the paper. To repair these label-level errors without losing samples unnecessarily, the lead author re-reviews every label that the Stage~7 cross-check flagged as suspicious, totaling 72 cases over 55 unique samples, comprising 48 verified-false claims (the labeled section did not contain the claim) and 24 unresolved labels (the labeled section header did not match any header in the gold paper). Each case receives one of five verdicts:
\begin{itemize}[leftmargin=*, itemsep=2pt, topsep=4pt]
\item \textbf{Repoint to parent} (23 cases), the label points at a fine-grained subsection that does not exist verbatim in the released pool, and the auditor supplies the parent section header that does.
\item \textbf{Repoint section} (22 cases), the labeled section header is wrong outright; the auditor supplies the correct header.
\item \textbf{Keep (rounding/precision ok)} (19 cases), the verbatim mismatch is innocuous (e.g., ``94.1\%'' label vs ``94.13\%'' in paper), no patch needed.
\item \textbf{Drop claim only} (1 case), one specific claim is retired but the sample survives with its remaining claims intact.
\item \textbf{Drop sample} (7 verdicts on 5 unique samples), the recall claim is fundamentally unrecoverable: the cited passage cannot be located in the released paper at any header, the gold paper's released body is missing entirely (only the appendix is present), or the terminal paper has no parsed body at all. The 5 retired samples (\texttt{mt\_0534}, \texttt{mt\_0650}, \texttt{mt\_0845}, \texttt{st\_0461}, \texttt{st\_0485}) cluster into three failure modes: \emph{claim cannot be located} (2 of 5), \emph{released paper pool missing the gold body} (2 of 5), and \emph{cited terminal paper has no parsed content} (1 of 5).
\end{itemize}
The 64 \emph{Repoint} and \emph{Keep} verdicts patch label headers in place; the released section-recall labels are post-triage. Only the 5 \emph{Drop sample} verdicts retire samples at this step.

\textbf{Step 2: Length filter (automated, $-27$ samples).} The remaining 1{,}038 samples are passed through a length filter that drops any sample whose gold-path papers (seed, bridge, or target) contain a single section longer than the p99.9 length cutoff of \textbf{70{,}015 characters}. These are HTML-parser collapse cases in which a parsing error has merged the main body and appendix into one giant section, making section-recall measurement meaningless. Twenty-seven samples meet this criterion, including cases where the labeled gold-evidence section is itself the parser-collapsed one, those samples are dropped because section-recall measurement on them would be ambiguous. The cutoff is calibrated so that legitimate long sections (e.g., dense methodology sections in foundation-model papers) are preserved.

\paragraph{Sample-difficulty distribution.} Figure~\ref{fig:difficulty_distribution} characterizes the benchmark's difficulty profile across three independent cuts, with each of the $1{,}011$ items grouped by how many of the $19$ evaluated models answered it correctly. The panel-19 coverage histogram (panel c) shows a smooth distribution centered on intermediate coverage: only $0.2\%$ of items are solved by every model and $1.1\%$ are failed by every model, with $44.8\%$ in the mid-coverage bucket ($7$--$12$ correct) and $31.6\%$ in the high-coverage bucket ($13$--$19$ correct). The joint distribution of question type and depth (panel a) gives a clear difficulty ordering: single-target depth-$1$ items are easiest ($58.9\%$ high-coverage), multi-target depth-$2$ items are hardest ($21.3\%$ high-coverage), and the intermediate cells (ST-d$2$, MT-d$1$) cluster at moderate difficulty, validating depth and target multiplicity as independent design axes. The consensus tier (panel b) acts as a separate construction-rigor signal: silver-tier samples (highest filter agreement during the audit) concentrate in high-coverage ($50.9\%$), while bronze-tier samples (most ambiguous during the audit) concentrate in low-coverage ($33.9\%$), reinforcing the tier as a measurement of construction confidence rather than a designed difficulty axis.

%% file: sections/Apdx_D.tex

\section{Experimental Configuration}
\label{app:setup}

\paragraph{Model naming convention.}
The main text refers to each evaluated model by its conventional release name (e.g., Gemini-3 Pro, Claude Opus 4.6). For reproducibility, Table~\ref{tab:model_ids} records the exact API model identifier or Hugging Face repository path invoked at evaluation time: closed-frontier models were called via their providers' production APIs at the snapshots indicated, and open-weight checkpoints were downloaded from the listed Hugging Face repositories.

\begin{table}[h]
\centering
\caption{Mapping from each display name used in the paper (and in Table~\ref{tab:main_results}) to the canonical API model identifier or Hugging Face repository path actually invoked. \texttt{Qwen3-30B-A3B} is the \texttt{Instruct-2507} snapshot. During our evaluation window (after the 2025-12-01 stable release of DeepSeek-V3.2~\citep{deepseekai2025deepseekv32pushingfrontieropen}), both DeepSeek API aliases resolve to the same underlying V3.2 checkpoint: \texttt{deepseek-chat} invokes the non-thinking mode and \texttt{deepseek-reasoner} the thinking mode. Reproducing our numbers therefore requires the V3.2 weights rather than the older V3/R1 weights to which these aliases pointed before the swap.}
\label{tab:model_ids}
\small
\setlength{\tabcolsep}{6pt}
\resizebox{\linewidth}{!}{%
\begin{tabular}{l l @{\qquad} l l}
\toprule
\textbf{Display name} & \textbf{Identifier} & \textbf{Display name} & \textbf{Identifier} \\
\midrule
Gemini-3 Pro       & \texttt{gemini-3-pro-preview}    & MiniMax-M2.7      & \texttt{MiniMaxAI/MiniMax-M2.7} \\
Claude Opus 4.6    & \texttt{claude-opus-4-6}         & Qwen3.5-27B       & \texttt{Qwen/Qwen3.5-27B} \\
GPT-5.3-codex      & \texttt{gpt-5.3-codex}           & Qwen3.5-35B-A3B   & \texttt{Qwen/Qwen3.5-35B-A3B} \\
Claude Sonnet 4.6  & \texttt{claude-sonnet-4-6}       & Qwen3.6-35B-A3B   & \texttt{Qwen/Qwen3.6-35B-A3B} \\
GPT-5.4            & \texttt{gpt-5.4}                 & Qwen3-32B         & \texttt{Qwen/Qwen3-32B} \\
Gemini-3 Flash     & \texttt{gemini-3-flash-preview}  & Qwen3-30B-A3B     & \texttt{Qwen/Qwen3-30B-A3B-Instruct-2507} \\
GLM-5.1            & \texttt{zai-org/GLM-5.1}         & Gemma-3-27B       & \texttt{google/gemma-3-27b-it} \\
DeepSeek-chat      & \texttt{deepseek-chat}           & Gemma-4-31B       & \texttt{google/gemma-4-31b-it} \\
DeepSeek-reasoner  & \texttt{deepseek-reasoner}       & Gemma-4-26B-A4B   & \texttt{google/gemma-4-26b-a4b-it} \\
Kimi-K2.5          & \texttt{moonshotai/kimi-k2.5}    & & \\
\bottomrule
\end{tabular}}
\end{table}

\paragraph{Backend choices.}
Closed-frontier models are accessed through the official APIs of their providers: Anthropic for Claude, OpenAI Responses API for GPT, and Google GenAI for Gemini. Open-frontier models are accessed via two API providers: \texttt{deepseek-chat} and \texttt{deepseek-reasoner} (both resolving to DeepSeek-V3.2~\citep{deepseekai2025deepseekv32pushingfrontieropen}; see Table~\ref{tab:model_ids} caption) through DeepSeek's own API, and Kimi-K2.5, MiniMax-M2.7, and GLM-5.1 through the OpenAI-compatible Together.ai endpoint. Compute-efficient open-weight models (Qwen3.x and Gemma~3/4 variants) are served locally with vLLM 0.19.1 using two-way tensor parallelism on a pair of H100 GPUs.

\paragraph{Sampling and reasoning configurations.}
All models receive identical system prompts and tool schemas. Sampling temperatures are set to each provider's defaults: 1.0 for Claude, 1.0 for GPT-5.x reasoning models, 0.6 with top-$p=0.95$ for Qwen and DeepSeek, and 1.0 for Gemini. Reasoning configurations are model-specific: Claude Opus 4.6 uses extended thinking with an 8{,}192-token budget combined with the external \texttt{think} tool; Claude Sonnet 4.6 uses \texttt{think} alone; GPT-5.3-codex and GPT-5.4 use \texttt{reasoning\_effort=medium}; MiniMax-M2.7 uses interleaved thinking; Qwen3.5/3.6 thinking variants use vLLM's \texttt{qwen3} reasoning parser with the \texttt{qwen3\_coder} tool-call parser to handle thinking blocks alongside tool calls.

\paragraph{Cost accounting.}
Per-model API cost and average per-sample token consumption for the 11 API-served models are reported in Table~\ref{tab:cost} for both the full-tool 19-model evaluation pass and the parallel zero-tool ablation; the 8 locally-served open-weight Qwen and Gemma checkpoints incur compute time on our local H100s but no API cost and are excluded. A single full-tool evaluation pass costs approximately \$1{,}294 across the 11 API models, ranging from \$11 (DeepSeek-chat) to \$406 (Claude Opus 4.6); the corresponding zero-tool ablation costs an additional \$259 across all 11 checkpoints. The two stability replicates (run-1 and run-2 on DeepSeek-chat and GPT-5.3 codex) add $\sim\$92$ each, and the no-search ablation on the same two checkpoints adds another $\sim\$90$. The data-generation pipeline (Stages 2--7 of Appendix~\ref{app:pipeline}), driven primarily by GPT-5.4 for question, distractor, and audit-prep generation alongside the three-model ensemble filter, accounts for an additional $\sim\$700$ in API cost. The aggregate API spend across the full project is approximately \$2.5K. Input-token consumption varies substantially with strategy: DeepSeek-chat averages 154K input tokens per sample reflecting sequential exploration, while GPT-5.3-codex averages 67K reflecting targeted retrieval. Reasoning-augmented models consume roughly $2\times$ the output tokens of non-reasoning models, indicating that the bottleneck on this task is action selection rather than answer generation.

\begin{table}[h]
\centering
\caption{Per-model API cost and average per-sample token consumption for the 11 API-served models on the 1{,}011-sample full split, alongside zero-tool ablation cost. All costs use the official per-token pricing for each provider as recorded in the released harness. The 8 open-weight Qwen and Gemma checkpoints (Qwen3-32B, Qwen3-30B-A3B, Qwen3.5-27B, Qwen3.5-35B-A3B, Qwen3.6-35B-A3B, Gemma-3-27B, Gemma-4-26B-A4B, Gemma-4-31B) are served locally on a pair of H100 GPUs and excluded.}
\label{tab:cost}
\small
\begin{tabular}{lrrrrr}
\toprule
Model & Full-tool (\$) & Zero-tool (\$) & \$/sample (full) & Input/sample (K) & Output/sample \\
\midrule
Gemini 3 Pro      & 159.27   & 37.14   & 0.158 & 87  & 5{,}003 \\
Gemini 3 Flash    & 77.49    & 25.17   & 0.077 & 169 & 8{,}517 \\
Claude Opus 4.6   & 405.54   & 79.47   & 0.401 & 107 & 3{,}890 \\
Claude Sonnet 4.6 & 205.23   & 43.86   & 0.203 & 92  & 4{,}165 \\
GPT-5.3-codex     & 80.32    & 14.80   & 0.079 & 67  & 2{,}208 \\
GPT-5.4           & 78.99    & 28.14   & 0.078 & 65  & 2{,}444 \\
DeepSeek-V3       & 11.00    & 0.80    & 0.011 & 154 & 2{,}818 \\
DeepSeek-R1       & 25.11    & 5.16    & 0.025 & 124 & 4{,}702 \\
GLM-5.1           & 155.66   & 12.56   & 0.154 & 98  & 3{,}734 \\
Kimi-K2.5         & 66.72    & 10.49   & 0.066 & 112 & 3{,}588 \\
MiniMax-M2.7      & 28.66    & 1.74    & 0.028 & 81  & 3{,}391 \\
\midrule
\textbf{Total}    & \textbf{1{,}293.99} & \textbf{259.33} & --- & --- & --- \\
\bottomrule
\end{tabular}
\end{table}

%% file: sections/Apdx_E.tex
\section{Diagnostic Detail and Robustness Checks}
\label{app:diagnostic_detail}

This appendix is organized as a section-by-section extension of Section~\ref{sec:experiments}: Section~\ref{app:per_axis} extends the per-axis decomposition of Section~\ref{subsec:main_results}; Section~\ref{app:trajectory_failure} extends the trajectory and failure analyses of Section~\ref{subsec:additional}; Section~\ref{app:robustness_validity} extends the robustness checks of Section~\ref{subsec:validity}; and Section~\ref{app:tool_ablations} extends the tool ablations of Section~\ref{subsec:ablations}.

\subsection{Per-axis decomposition (extends Section~\ref{subsec:main_results})}
\label{app:per_axis}

\paragraph{Per-model stopping behavior across three scopes.}
\label{app:mt_diag}
Section~\ref{subsec:main_results} reports four per-family stopping signatures derived from what each model does after reaching a gold paper. Table~\ref{tab:mt_diag_full} gives the full-panel decomposition across three scopes: \emph{MT-fail} (multi-target items on which paper recall failed; categorized by the post-first-gold action), \emph{ST} (single-target items, action after the gold read), and \emph{MT-both} (multi-target items where both gold papers were reached; action after the second gold read). MT-fail uses four mutually exclusive categories: \textbf{0-Gold} (trajectory submitted without reaching any gold paper), \textbf{Prem} (reached a gold but made no further read or navigation tool call, premature commit), \textbf{Explored} (further navigation or non-gold reads, tried but missed the second gold), \textbf{Stayed} (further reads only on the already-reached gold). ST and MT-both each use three categories: \textbf{Imm} (submit immediately), \textbf{Thnk} (think before submit), \textbf{Read} (read additional sections before submit); a small residual ($\leq 25$\%) navigates without reading and is omitted.

\begin{table}[h]
\centering
\caption{Stopping behavior across MT-fail, ST, and MT-both scopes for the 19-model panel. Each block's $n$ column reports trajectories in scope; subsequent columns report the within-scope share (\%) of each category. MT-fail's four categories sum to 100\%; ST and MT-both's three categories sum to $\leq 100$\% (residual is navigation without reading). Scope membership counts only reads that returned content, as defined in Section~\ref{subsec:axes}; the post-gold action categories classify tool calls, that is, behavior, regardless of delivery.}
\label{tab:mt_diag_full}
\scriptsize
\setlength{\tabcolsep}{3pt}
\renewcommand{\arraystretch}{0.95}
\begin{tabular}{l rrrrr rrrr rrrr}
\toprule
 & \multicolumn{5}{c}{\textbf{MT-fail}} & \multicolumn{4}{c}{\textbf{ST (after gold)}} & \multicolumn{4}{c}{\textbf{MT-both (after both)}} \\
\cmidrule(lr){2-6} \cmidrule(lr){7-10} \cmidrule(lr){11-14}
Model & $n$ & 0-Gold & Prem & Explored & Stayed & $n$ & Imm & Thnk & Read & $n$ & Imm & Thnk & Read \\
\midrule
\multicolumn{14}{l}{\textit{Closed Frontier}} \\
Gemini-3 Pro          & 110 & 20.9 & 17.3 & 40.0 & 21.8 & 493 & 35.7 & 11.6 & 48.7 & 333 & 39.6 & 10.2 & 47.4 \\
Gemini-3 Flash        & 221 & 24.0 &  4.1 & 56.6 & 15.4 & 411 &  5.1 &  1.5 & 89.1 & 222 & 13.1 &  3.6 & 74.3 \\
GPT-5.3 codex         & 167 & 16.2 & 34.7 & 31.1 & 18.0 & 461 & 47.3 &  0.2 & 47.1 & 276 & 66.3 &  1.1 & 22.5 \\
GPT-5.4               & 158 & 17.1 & 53.2 & 27.2 &  2.5 & 459 & 31.4 & 14.4 & 45.8 & 285 & 38.9 & 29.5 & 21.1 \\
Claude Opus 4.6       &  99 & 33.3 & 37.4 & 13.1 & 16.2 & 496 &  0.4 & 67.7 & 31.7 & 344 &  2.0 & 64.8 & 33.1 \\
Claude Sonnet 4.6     & 177 & 51.4 & 25.4 & 13.0 & 10.2 & 453 &  0.4 & 70.4 & 29.1 & 266 &  0.4 & 64.3 & 34.6 \\
\midrule
\multicolumn{14}{l}{\textit{Open Frontier}} \\
GLM-5.1               & 145 & 26.2 & 33.1 & 29.0 & 11.7 & 436 &  3.0 & 50.5 & 46.1 & 298 &  4.0 & 39.9 & 55.4 \\
DeepSeek-reasoner     & 196 & 18.9 &  3.6 & 66.8 & 10.7 & 409 & 10.0 &  1.0 & 86.8 & 247 & 20.2 &  2.0 & 73.3 \\
DeepSeek-chat         & 167 & 18.0 &  1.8 & 64.1 & 16.2 & 439 &  1.4 &  3.0 & 92.7 & 276 &  4.7 &  9.4 & 77.2 \\
Kimi-K2.5             & 191 & 16.8 & 11.0 & 57.6 & 14.7 & 434 &  3.5 & 16.6 & 76.7 & 252 &  5.6 & 20.2 & 69.4 \\
MiniMax-M2.7          & 198 & 26.8 & 17.7 & 40.9 & 14.6 & 379 &  5.0 & 28.8 & 64.4 & 245 &  7.3 & 29.0 & 62.9 \\
\midrule
\multicolumn{14}{l}{\textit{Compute-Efficient}} \\
Gemma-4-31B           & 212 & 34.4 & 12.7 & 31.6 & 21.2 & 353 & 45.0 &  9.6 & 45.3 & 231 & 13.0 & 38.1 & 48.9 \\
Gemma-4-26B-A4B       & 373 & 66.2 &  7.0 &  9.1 & 17.7 & 200 & 14.0 &  6.5 & 78.0 &  70 & 32.9 & 18.6 & 48.6 \\
Gemma-3-27B           & 419 & 68.5 & 20.0 &  6.2 &  5.3 &  79 & 78.5 &  0.0 & 21.5 &  24 & 62.5 &  0.0 & 37.5 \\
Qwen3.5-27B           & 182 & 17.6 & 11.0 & 54.4 & 17.0 & 385 & 28.1 &  7.0 & 63.6 & 261 & 29.5 & 18.0 & 52.1 \\
Qwen3.6-35B-A3B       & 309 & 45.3 &  7.4 & 30.1 & 17.2 & 287 & 12.9 & 16.4 & 69.7 & 134 & 17.9 & 29.9 & 52.2 \\
Qwen3.5-35B-A3B       & 234 & 28.2 &  3.8 & 48.7 & 19.2 & 327 & 10.7 &  5.2 & 82.9 & 209 & 14.4 & 11.0 & 72.7 \\
Qwen3-30B-A3B         & 239 & 49.8 &  6.3 & 27.6 & 16.3 & 267 &  3.7 & 34.8 & 61.4 & 204 &  5.4 & 65.7 & 28.4 \\
Qwen3-32B             & 403 & 91.1 &  5.2 &  3.0 &  0.7 &  87 & 77.0 &  0.0 & 21.8 &  40 & 60.0 & 12.5 & 27.5 \\
\bottomrule
\end{tabular}
\end{table}

\paragraph{Per-cell decomposition for representative models.}
\label{app:cell_decomp}
Crossing the question-type axis (single-target vs.\ multi-target) with the depth axis (depth-1 vs.\ depth-2) yields a $2 \times 2$ grid that exposes per-model discrimination patterns. Table~\ref{tab:cell_decomp} reports per-cell accuracy for one representative checkpoint from each tier; spreads across the four cells diagnose whether a model is bottlenecked by synthesis depth or by tool-environment overhead.

\begin{table}[h]
\centering
\caption{Per-cell accuracy on the question-type $\times$ depth grid for one representative checkpoint from each tier (point estimate $\pm$ Wilson 95\% CI half-width). $\Delta$ is the drop from the easiest to hardest cell. GPT-5.3 codex holds within $17$\,pp across all four cells; DeepSeek-chat collapses on multi-target depth-2; Gemma-4-31B is uniformly capped, suggesting that what limits its accuracy is tool-environment overhead rather than synthesis depth.}
\label{tab:cell_decomp}
\small
\setlength{\tabcolsep}{8pt}
\begin{tabular}{lcccc r}
\toprule
Model & st\_d1 & st\_d2 & mt\_d1 & mt\_d2 & $\Delta_{\text{st\_d1}-\text{mt\_d2}}$ \\
 & ($n=90$) & ($n=478$) & ($n=396$) & ($n=47$) & (pp) \\
\midrule
GPT-5.3 codex     & $94.4 \pm 3.2$ & $78.0 \pm 3.5$ & $85.4 \pm 3.1$ & $80.9 \pm 8.7$ & $+13.6$ \\
DeepSeek-chat& $73.3 \pm 8.0$ & $53.8 \pm 4.4$ & $56.6 \pm 4.8$ & $48.9 \pm 13.8$ & $+24.4$ \\
Gemma-4-31B       & $72.2 \pm 8.2$ & $56.3 \pm 4.4$ & $72.7 \pm 4.2$ & $70.2 \pm 11.1$ &  $+2.0$ \\
\bottomrule
\end{tabular}
\end{table}

\paragraph{Reasoning-type breakdown.}
\label{app:reasoning_type}
Each AgentHop item is tagged at construction with one of four reasoning-type labels. \textbf{GROUND} questions ask for a paper-grounded factual claim (n$=330$); \textbf{METHOD} questions ask about a specific technical procedure used by a paper (n$=124$); \textbf{MOTIVE} questions ask why a paper was written or how it relates to prior work (n$=265$); \textbf{RESULT} questions ask about the empirical outcomes a paper reports (n$=292$). The taxonomy was applied at Stage 3 of the construction pipeline of Appendix~\ref{app:pipeline} by the Q/A-generation model and verified during the Stage 7 audit; reasoning-type labels are not used to score answers and do not enter the four-axis decomposition of Section~\ref{subsec:axes}.

Across the 19-model panel, mean accuracy by reasoning type ranks METHOD ($57.1\%$) > RESULT ($54.5\%$) > GROUND ($54.1\%$) > MOTIVE ($53.1\%$): METHOD questions, which point to specific technical procedures often documented in named sections (\emph{Architecture}, \emph{Method}), are the easiest reasoning type for the panel; MOTIVE questions, which require integrating across the introduction and related-work content of one or more papers to reconstruct authorial intent, are the hardest. The spread is small ($\sim 4$\,pp), so reasoning type is too weak to support a main-body axis. The per-model breakdown does carry one diagnostic signal worth flagging: Claude Sonnet drops $12$\,pp from its METHOD accuracy ($83.9\%$) to its MOTIVE accuracy ($71.7\%$), the largest within-model spread in the panel and the clearest example of a model whose synthesis-heavy reasoning sub-type is bottlenecked relative to its lookup-style reasoning.

\begin{figure}[h]
\centering
\includegraphics[width=0.72\textwidth]{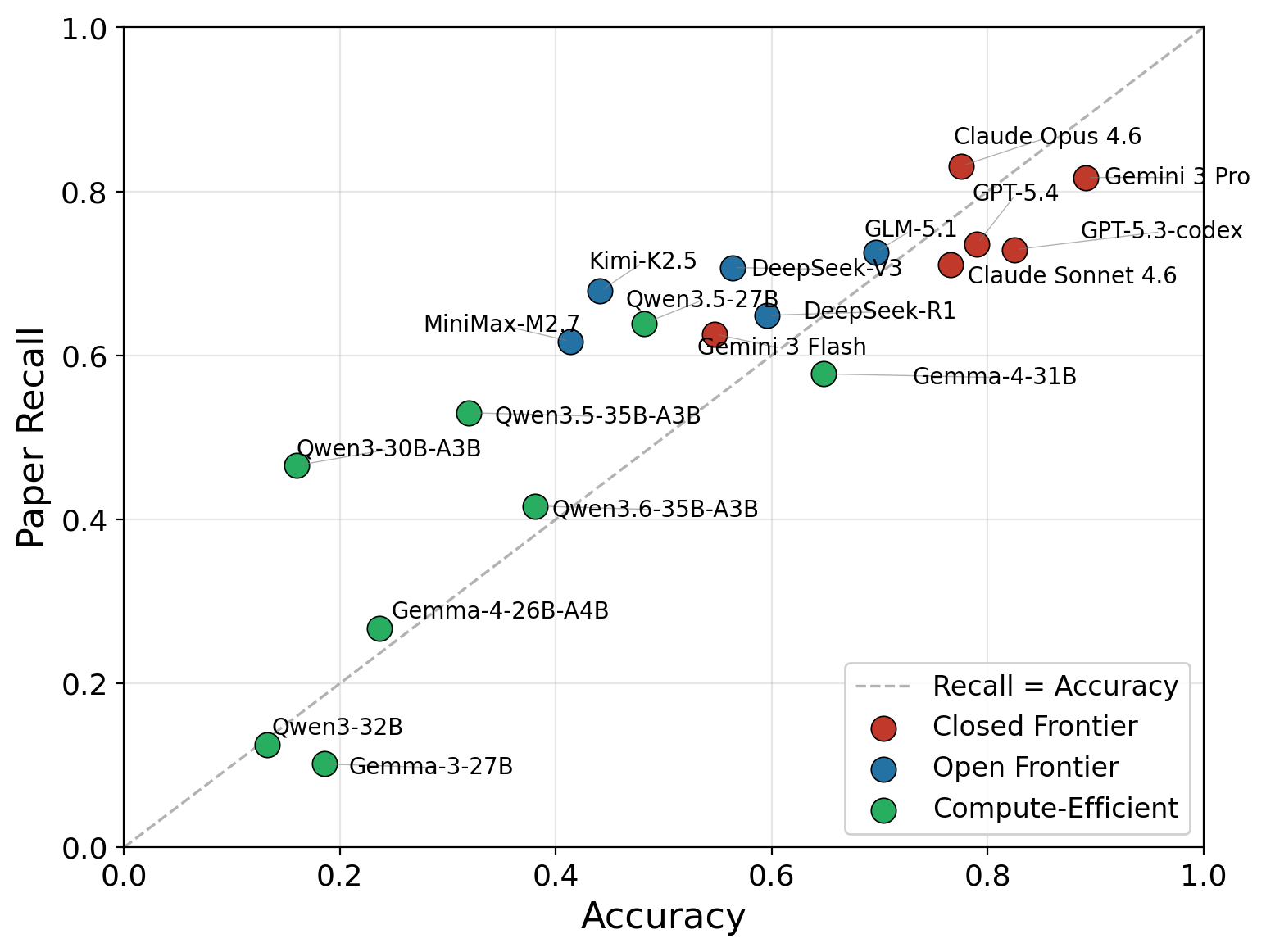}
\caption{Accuracy versus paper recall across the 19-model panel, colored by tier. The diagonal $y=x$ marks accuracy tracking retrieval one-to-one. Models above the diagonal (Gemini-3 Pro, GPT-5.3 codex, Gemma-4-31B) combine retrieval with synthesis lift; models below (Kimi-K2.5, MiniMax-M2.7, the Qwen MoE family, DeepSeek-chat) retrieve correctly but fail to convert.}
\label{fig:accuracy_vs_recall}
\end{figure}

\paragraph{Joint view: accuracy vs.\ paper recall.}
Figure~\ref{fig:accuracy_vs_recall} plots accuracy against paper recall for all 19 models. Most checkpoints cluster near the diagonal $y=x$, indicating that retrieval success is the dominant driver of accuracy. Two diagnostic profiles diverge from this regime: models above the diagonal accumulate accuracy beyond what retrieval alone explains, while models below retrieve correctly but lose accuracy at the synthesis step.

\begin{table}[h]
\centering
\caption{Evidence-contact decomposition of the $11{,}072$ recall-credited trajectories,
pooled over the 19-model panel. Conversion is cap-corrected. The zero-tool column reports the
same models' closed-book accuracy on the same items.}
\label{tab:evidence_contact}
\small
\begin{tabular}{lrrrr}
\toprule
Group & $n$ & Share & Conversion & Zero-tool \\
\midrule
Read a \texttt{direct}-labelled section (header match) & 4{,}725 & 42.7\% & 74.0\% & 35.2\% \\
Read the labelled content inside its parent section    & 4{,}765 & 43.0\% & 78.1\% & 40.1\% \\
Read no \texttt{direct}-labelled content               & 1{,}582 & 14.3\% & 49.3\% & 31.5\% \\
\bottomrule
\end{tabular}
\end{table}

\paragraph{Section-recall label mapping.}
\label{app:section_recall_mapping}
Audit labels cite evidence at each paper's native granularity. Across the $2{,}773$ unique \texttt{direct} labels, $53.0$\% name a top-level section, $46.8$\% name a subsection, and five name citation targets such as tables; the five are repaired in the release. The sandbox's readable unit is the top-level section, so a subsection label is credited when its header appears verbatim in the text of a delivered section. This is the same resolution rule the
audit pipeline applies during label verification. Table~\ref{tab:evidence_contact} decomposes the $11{,}072$ recall-credited trajectories by evidence contact: $85.7$\% verifiably received \texttt{direct}-labelled evidence text, and conversion separates the groups by roughly $25$ points, which validates the labels as predictors of synthesis success.

\begin{table}[h]
\centering
\caption{Exact McNemar tests on common recalled items (discordant pairs; two-sided $p$).
Top: adjacent pairs among the six highest multi-target conversion scores. Bottom: tier
conversion leader against tier runner-up, per axis.}
\label{tab:mcnemar}
\small
\begin{tabular}{llrrr}
\toprule
 & Pair & $n$ & Discordant & $p$ \\
\midrule
MT & Gemini-3 Pro vs Claude Sonnet 4.6 & 226 & 20/3 & 0.0005 \\
MT & Claude Sonnet 4.6 vs GPT-5.4 & 193 & 24/16 & 0.27 \\
MT & GPT-5.4 vs GLM-5.1 & 227 & 36/22 & 0.09 \\
MT & GLM-5.1 vs DeepSeek-reasoner & 203 & 35/21 & 0.08 \\
MT & DeepSeek-reasoner vs MiniMax-M2.7 & 168 & 42/12 & 0.0001 \\
\midrule
Closed, ST & Gemini-3 Pro vs Claude Sonnet 4.6 & 437 & 25/13 & 0.073 \\
Closed, MT & Gemini-3 Pro vs Claude Sonnet 4.6 & 226 & 20/3 & 0.0005 \\
Open, ST & GLM-5.1 vs DeepSeek-reasoner & 368 & 56/28 & 0.0030 \\
Open, MT & GLM-5.1 vs DeepSeek-reasoner & 203 & 35/21 & 0.081 \\
Comp.-eff., ST & Gemma-4-31B vs Qwen3.5-27B & 315 & 56/30 & 0.0067 \\
Comp.-eff., MT & Gemma-4-31B vs Qwen3.6-35B-A3B & 98 & 22/5 & 0.0015 \\
\bottomrule
\end{tabular}
\end{table}

\paragraph{Paired conversion tests.}
\label{app:paired_tests}
Marginal confidence intervals on different item sets confound model ability with item difficulty. We therefore compare conversion with exact McNemar tests on common recalled items, cap-corrected. The top panel walks down the six highest multi-target conversion scores; the bottom panel tests each tier's conversion leader against its runner-up on both axes. Table~2 shades a conversion tier leader only where the corresponding test rejects at $p<0.05$.

\subsection{Trajectory and failure analysis (extends Section~\ref{subsec:additional})}
\label{app:trajectory_failure}

\paragraph{Trajectory deeper-dive.}
\label{app:trajectory}
This subsection expands the strategy analysis of Section~\ref{subsec:additional} with three lenses: bigram-level accuracy associations, per-family trigram fingerprints, the early-submit-after-wrong-paper signature, and per-model tool-call transition visualisations. Each lens isolates a different trajectory-shape signal that aggregate accuracy does not surface.

\textbf{Bigram accuracy associations across all 19 models.} For each ordered pair of tool-call actions $(a_i, a_{i+1})$ we compute the marginal accuracy difference between trajectories that contain the bigram and those that do not, applying a Bonferroni correction over the 49 candidate bigrams ($7\times 7$ tools). Table~\ref{tab:bigram_stats} reports the top productive and anti-productive bigrams ranked by panel-mean $|\Delta|$, restricted to bigrams that appear in at least $10$ of $19$ models so that the panel mean is well-defined.

\begin{table}[h]
\centering
\caption{Top productive and anti-productive bigrams across the 19-model panel, ranked by panel-mean $|\Delta|$. $\Delta$ is the panel-mean accuracy difference between trajectories containing and not containing the bigram. \emph{Sig. count} is the number of individual models for which the per-model effect remains significant after Bonferroni correction over the $7\times 7$ candidate space; \emph{Models} is the number with the bigram present in $\geq 1$ trajectory.}
\label{tab:bigram_stats}
\small
\setlength{\tabcolsep}{6pt}
\begin{tabular}{l r r r}
\toprule
Bigram & $\Delta$ (pp) & Sig.\ count & Models \\
\midrule
\multicolumn{4}{l}{\textit{Productive (positive $\Delta$)}} \\
\texttt{read\_section} $\to$ \texttt{think}             & $+13.6$ & $7/19$ & $14/19$ \\
\texttt{think} $\to$ \texttt{submit\_answer}            & $+12.6$ & $6/19$ & $14/19$ \\
\texttt{list\_sections} $\to$ \texttt{read\_section}    & $+8.4$  & $2/19$ & $13/19$ \\
\texttt{get\_references} $\to$ \texttt{get\_paper\_info}& $+8.3$  & $5/19$ & $15/19$ \\
\texttt{get\_paper\_info} $\to$ \texttt{get\_references}& $+8.2$  & $4/19$ & $14/19$ \\
\texttt{get\_paper\_info} $\to$ \texttt{list\_sections} & $+7.3$  & $3/19$ & $15/19$ \\
\texttt{read\_section} $\to$ \texttt{submit\_answer}    & $+5.3$  & $7/19$ & $15/19$ \\
\midrule
\multicolumn{4}{l}{\textit{Anti-productive (negative $\Delta$)}} \\
\texttt{get\_paper\_info} $\to$ \texttt{submit\_answer} & $-14.4$ & $3/19$ & $13/19$ \\
\texttt{get\_references} $\to$ \texttt{read\_section}   & $-11.7$ & $1/19$ & $14/19$ \\
\texttt{read\_section} $\to$ \texttt{read\_section}     & $-11.5$ & $7/19$ & $15/19$ \\
\texttt{get\_references} $\to$ \texttt{submit\_answer}  & $-10.8$ & $1/19$ & $10/19$ \\
\texttt{list\_sections} $\to$ \texttt{submit\_answer}   & $-10.6$ & $0/19$ & $11/19$ \\
\texttt{get\_paper\_info} $\to$ \texttt{read\_section}  &  $-9.9$ & $2/19$ & $15/19$ \\
\bottomrule
\end{tabular}
\end{table}

\emph{Interpretation.} Three families of habit emerge. The \emph{deliberation} pair (\texttt{read\_section}$\to$\texttt{think}, \texttt{think}$\to$\texttt{submit\_answer}) reproduces the action--observation--think regime of \citet{yao2023reactsynergizingreasoningacting} and is uniformly productive on AgentHop. The \emph{scaffolded-exploration} cluster (\texttt{list\_sections}$\to$\texttt{read\_section}, \texttt{get\_references}$\to$\texttt{get\_paper\_info}, \texttt{get\_paper\_info}$\to$\texttt{get\_references}, \texttt{get\_paper\_info}$\to$\texttt{list\_sections}) carries +7 to +8\,pp regardless of which specific transition is involved, indicating that the productive signal is structural rather than tied to any single tool: any sequence that pauses to discover before committing pays off. The \emph{anti-productive} cluster splits into two distinct pathologies. Three of the negative bigrams (\texttt{get\_paper\_info}$\to$\texttt{submit\_answer}, \texttt{get\_references}$\to$\texttt{submit\_answer}, \texttt{list\_sections}$\to$\texttt{submit\_answer}) capture an \emph{abandon-after-metadata} pattern in which the model commits without reading any content; these are systematic --10 to --15\,pp losses. Two more (\texttt{read\_section}$\to$\texttt{read\_section}, \texttt{get\_references}$\to$\texttt{read\_section}, \texttt{get\_paper\_info}$\to$\texttt{read\_section}) capture a \emph{shortcut-read} pattern that skips deliberation or scaffolding before content access. AgentHop therefore rewards models that pace actions through deliberation or scaffolded exploration and punishes models that grab actions impulsively or abandon prematurely; the same conclusion is invisible from accuracy alone but reads directly off the bigram table.

\textbf{Per-family trigram fingerprints.} A complementary lens looks at length-3 sub-patterns: which trigram is each model's most-frequent triple of consecutive tool calls, and how concentrated is that pattern. Table~\ref{tab:trigram_top1} reports each model's top-1 trigram and the fraction of all observed trigrams it accounts for, grouped by tier. Tool abbreviations: gpi = \texttt{get\_paper\_info}, gr = \texttt{get\_references}, ls = \texttt{list\_sections}, rs = \texttt{read\_section}, th = \texttt{think}, sub = \texttt{submit\_answer}.

\begin{table}[h]
\centering
\caption{Per-model top-1 trigram and concentration share. \emph{Concentration} is the fraction of all length-3 sub-patterns occupied by the most-frequent trigram (higher = more strategically locked-in). Models grouped by tier; ranked within tier by concentration.}
\label{tab:trigram_top1}
\small
\setlength{\tabcolsep}{6pt}
\begin{tabular}{l l r}
\toprule
Model & Top-1 trigram & Concentration \\
\midrule
\multicolumn{3}{l}{\textit{Closed Frontier}} \\
Gemini-3 Flash    & gpi$\to$ls$\to$rs   & $0.100$ \\
Gemini-3 Pro      & ls$\to$rs$\to$rs    & $0.091$ \\
Claude Sonnet 4.6 & gpi$\to$gr$\to$th   & $0.090$ \\
GPT-5.4           & ls$\to$rs$\to$rs    & $0.089$ \\
GPT-5.3 codex     & gpi$\to$ls$\to$rs   & $0.089$ \\
Claude Opus 4.6   & rs$\to$th$\to$sub   & $0.084$ \\
\midrule
\multicolumn{3}{l}{\textit{Open Frontier}} \\
Kimi-K2.5         & ls$\to$rs$\to$rs    & $0.087$ \\
MiniMax-M2.7      & ls$\to$rs$\to$rs    & $0.079$ \\
DeepSeek-reasoner & gpi$\to$ls$\to$rs   & $0.079$ \\
GLM-5.1           & ls$\to$rs$\to$rs    & $0.072$ \\
DeepSeek-chat     & ls$\to$rs$\to$rs    & $0.071$ \\
\midrule
\multicolumn{3}{l}{\textit{Compute-Efficient}} \\
Gemma-3-27B       & gpi$\to$gr$\to$gpi  & $0.196$ \\
Qwen3.5-27B       & gpi$\to$ls$\to$rs   & $0.144$ \\
Qwen3-32B         & gpi$\to$ls$\to$rs   & $0.135$ \\
Qwen3-30B-A3B     & gpi$\to$gr$\to$gpi  & $0.131$ \\
Gemma-4-26B-A4B   & rs$\to$rs$\to$rs    & $0.125$ \\
Qwen3.6-35B-A3B   & gpi$\to$ls$\to$rs   & $0.120$ \\
Gemma-4-31B       & ls$\to$rs$\to$rs    & $0.102$ \\
Qwen3.5-35B-A3B   & gpi$\to$ls$\to$rs   & $0.093$ \\
\bottomrule
\end{tabular}
\end{table}

\emph{Interpretation.} Strategic diversity decreases as one moves down the tier hierarchy, and the failure mode at the bottom is concrete. Closed-frontier checkpoints invent five different top-1 trigrams across six models, with Claude Opus 4.6 the only model whose most-frequent trigram is the canonical ReAct chain itself (\texttt{rs}$\to$\texttt{th}$\to$\texttt{sub}); Sonnet's \texttt{gpi}$\to$\texttt{gr}$\to$\texttt{th} is the only think-terminating trigram outside Opus, consistent with the verify-before-commit family signature reported in Section~\ref{subsec:main_results}. Open-frontier checkpoints converge: 4 of 5 share the greedy \texttt{ls}$\to$\texttt{rs}$\to$\texttt{rs} pattern as top-1, suggesting less differentiated post-training across this group. Compute-efficient checkpoints are bimodal. Five of eight (the four Qwen variants plus Gemma-4-31B) reach the productive \texttt{gpi}$\to$\texttt{ls}$\to$\texttt{rs} or \texttt{ls}$\to$\texttt{rs}$\to$\texttt{rs} pattern at concentrations comparable to the closed tier. Three lock onto pathologies: Gemma-3-27B ($19.6\%$) and Qwen3-30B-A3B ($13.1\%$) concentrate on a metadata-only loop \texttt{gpi}$\to$\texttt{gr}$\to$\texttt{gpi} that never enters \texttt{read\_section}, and Gemma-4-26B-A4B locks onto \texttt{rs}$\to$\texttt{rs}$\to$\texttt{rs} greedy reads ($12.5\%$). The two patterns map directly to the Section~\ref{subsec:main_results} resource pathologies: metadata-only loops drive the under-engagement profile (Gemma-3-27B, Qwen3-32B floor accuracy), and greedy reads drive the token-cap exhaustion of Gemma-4-26B-A4B. Trigram concentration thus distinguishes \emph{which kind} of failure a low-accuracy model is exhibiting, not just \emph{whether} it is failing.

\textbf{Early-submit-after-wrong-paper signature.} A separate trajectory-level pattern is submitting an answer within three actions of reading a non-gold paper, which associates with $\Delta = -21.1$\,pp accuracy on average across the panel and is significant in $12$ of $19$ models. The per-instance penalty is largest in checkpoints for which the behavior is atypical, Claude Sonnet, Claude Opus, Gemma-4-31B, and GLM-5.1 each lose at least forty percentage points on the trajectories where they make this mistake. In the compute-efficient checkpoints where the behavior is endemic (roughly a quarter to a third of all trajectories on Qwen3.5-27B, Gemma-4-31B, and Qwen3.6-35B-A3B), the per-instance penalty shrinks but absolute outcomes stay poor: the bottleneck has migrated from this single bigram to broader navigation failure.

\begin{figure}[h]
\centering
\includegraphics[width=\textwidth]{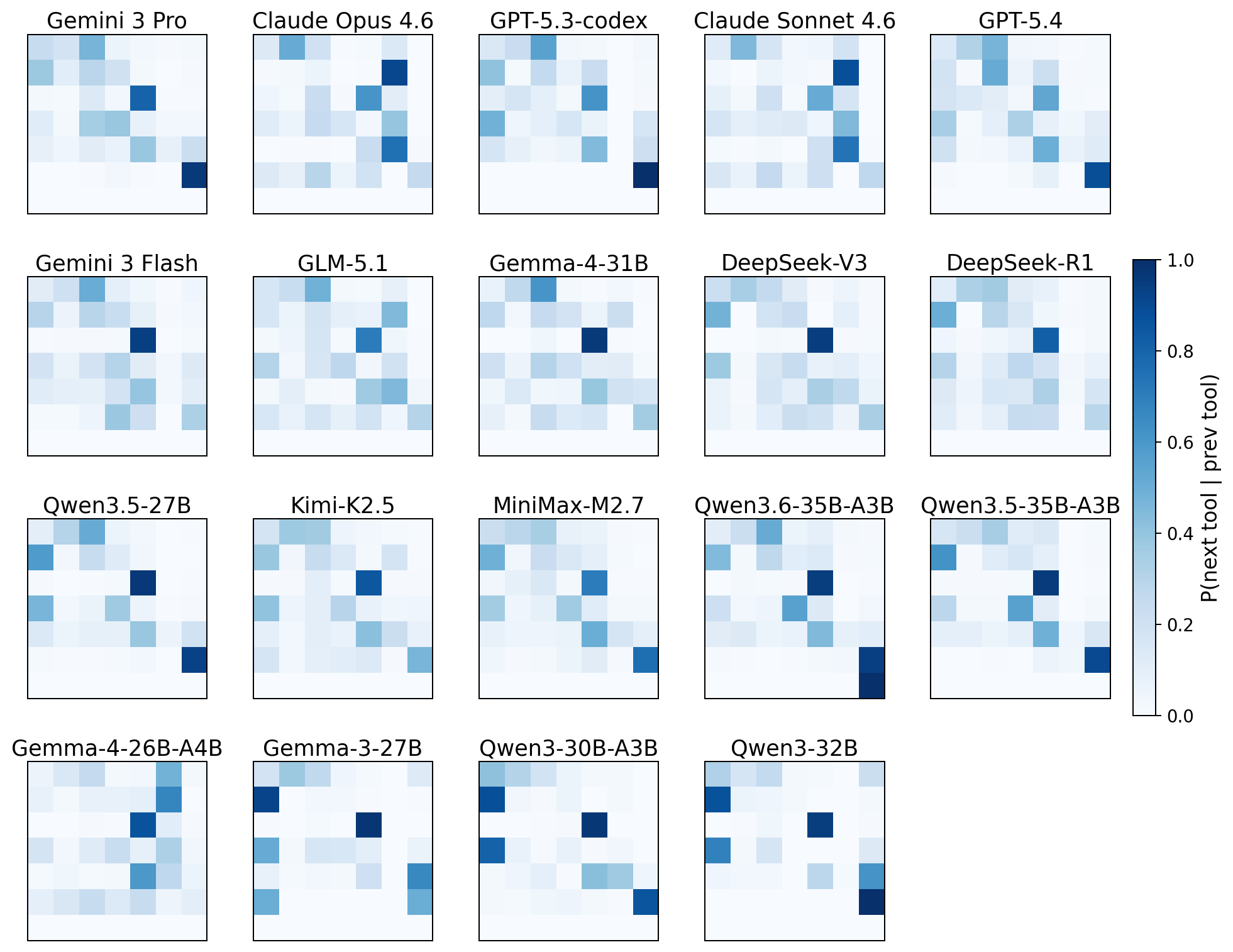}
\caption{Per-model tool-call transition matrices for the 19 evaluated models. Each panel is a $7\times 7$ heatmap whose rows index the previous tool action and whose columns index the next; cell color encodes the row-normalised conditional probability $P(\text{next} \mid \text{prev})$ (deeper blue = higher transition probability). Within every panel the tool order along both axes is, top-to-bottom and left-to-right: \texttt{get\_paper\_info}, \texttt{get\_references}, \texttt{list\_sections}, \texttt{search\_papers}, \texttt{read\_section}, \texttt{think}, \texttt{submit\_answer}. Panels are ordered to keep within-family checkpoints adjacent.}
\label{fig:transition_matrices}
\end{figure}

\textbf{Per-model tool-call transition matrices.} Figure~\ref{fig:transition_matrices} visualises the $7\times 7$ tool-call transition probabilities per model. Five visually distinct strategy archetypes are visible across the 19-checkpoint panel. The \emph{Anthropic ReAct band}: Claude Opus 4.6 and Sonnet 4.6 show a strong \texttt{read\_section}$\to$\texttt{think} transition with darker bands originating from \texttt{think}, indicating deliberate post-reading reflection. The \emph{OpenAI section-list dominance}: GPT-5.3 codex and GPT-5.4 emphasize \texttt{list\_sections}$\to$\texttt{read\_section}, with section-listing as the canonical bridge between metadata and content. The \emph{synthesis-bottlenecked greedy chain}: DeepSeek-chat and Kimi-K2.5 show long bands of \texttt{read\_section}$\to$\texttt{read\_section} chains without intervening think, the visual signature of the synthesis-bottlenecked profile. The \emph{metadata-only loop pathology}: Gemma-3-27B and Qwen3-30B-A3B concentrate transitions on \texttt{get\_paper\_info}$\to$\texttt{get\_references}$\to$\texttt{get\_paper\_info}, with no \texttt{read\_section} band at all. The \emph{diffuse exploration} archetype: Gemini-3 Flash, GLM-5.1, MiniMax-M2.7, and the Qwen MoE checkpoints show heatmaps without strong dominant bands, consistent with their broader strategic inconsistency and lower top-1 trigram concentrations in Table~\ref{tab:trigram_top1}.

\emph{Interpretation.} The same five archetypes recur across all $1{,}011$ samples per model, indicating that trajectory shape is largely a property of training composition rather than per-sample task content. Two stronger-form claims follow. First, the archetypes align cleanly with the per-family stopping signatures of Section~\ref{subsec:main_results} ¶1: ReAct band $\leftrightarrow$ verify-before-commit, section-list dominance $\leftrightarrow$ confident-commit, greedy chains $\leftrightarrow$ over-search. Second, the metadata-only loop archetype is observed only in legacy compute-efficient checkpoints whose pre-training predates broad agentic post-training, suggesting that the agent-loop competence is acquired during a specific phase of post-training that some lineages have undergone and others have not. Tool-call transition shape is therefore a family fingerprint that is stable per model and predictive of failure mode at the panel level; as with the signatures in Section~\ref{subsec:main_results}, it is measured through each provider's serving interface and characterizes the model-plus-interface system.

\paragraph{Non-submission terminations across the 19-model pool.}
\label{app:na_decomp_full}
Table~\ref{tab:na_decomp_full} extends the resource-axis analysis of Section~\ref{subsec:main_results} with the per-model breakdown of non-submission terminations across the three caps. Three failure modes recur, and each clusters by architecture rather than by capability tier. \emph{Token overflow} is the dominant terminator on compute-efficient sparse-active checkpoints, Gemma-4-26B-A4B alone exhausts the token cap on more than a third of trajectories, and also drives the Gemini-3 Flash and DeepSeek-chat non-submission rates. \emph{Max-turns} terminations cluster in the Qwen MoE and Gemma sparse-active families, where the model continues to issue cheap tool calls without ever committing. \emph{Budget exhaustion} appears almost exclusively in checkpoints with parallel tool-call regimes (GPT-5.4, MiniMax-M2.7, GLM-5.1), which drain the 30-point pool faster than turns or tokens accumulate. The pattern reinforces the main-text observation that the binding resource constraint is itself a model signature: the pairing of architecture, reasoning channel, and serving environment determines which of the three caps a model meets first.

\begin{table}[h]
\centering
\caption{Non-submission terminations across the 19-model panel under the cap-corrected protocol. NA is the fraction of trajectories that fail to produce a parseable answer letter; the three rightmost columns decompose NA by terminating cap (max-turns, budget, token-limit; \% of all trajectories). A small residual ($\leq 0.7$\,pp on Qwen3-32B and Qwen3-30B-A3B) reflects voluntary \texttt{submit\_answer} calls with non-letter output. Termination categories are defined in Appendix~\ref{app:error_handling}.}
\label{tab:na_decomp_full}
\scriptsize
\setlength{\tabcolsep}{4pt}
\renewcommand{\arraystretch}{0.95}
\begin{tabular}{l rrrrr @{\hskip 1.5em} l rrrrr}
\toprule
Model & Acc & NA & MaxT & Bud & Tok & Model & Acc & NA & MaxT & Bud & Tok \\
\midrule
\multicolumn{6}{l}{\textit{Closed Frontier}}                & \multicolumn{6}{l}{\textit{Open Frontier}} \\
Gemini-3 Pro      & 89.1 &  4.5 & 0.0 & 0.0 &  4.5 & GLM-5.1           & 69.6 &  7.9 & 0.4 & 3.2 &  4.4 \\
GPT-5.3 codex     & 82.5 &  1.0 & 0.0 & 0.7 &  0.3 & DeepSeek-R1       & 59.6 &  6.4 & 0.3 & 0.0 &  6.1 \\
GPT-5.4           & 79.0 &  5.2 & 0.0 & 4.9 &  0.3 & DeepSeek-V3       & 56.4 & 14.3 & 0.3 & 0.0 & 14.0 \\
Claude Opus 4.6   & 77.5 &  6.7 & 0.0 & 0.0 &  6.7 & Kimi-K2.5         & 44.1 &  9.9 & 2.0 & 2.4 &  5.5 \\
Claude Sonnet 4.6 & 76.6 &  4.4 & 0.6 & 0.0 &  3.8 & MiniMax-M2.7      & 41.3 &  7.2 & 1.3 & 4.8 &  1.1 \\
Gemini-3 Flash    & 54.7 & 31.1 & 0.4 & 0.2 & 30.4 &                   &      &      &     &     &      \\
\midrule
\multicolumn{12}{l}{\textit{Compute-Efficient}} \\
Gemma-4-31B     & 64.8 &  2.5 & 0.1 & 0.0 &  2.4 & Gemma-4-26B-A4B & 23.6 & 44.2 & 5.7 & 0.3 & 38.2 \\
Qwen3.5-27B     & 48.2 &  9.8 & 2.1 & 0.6 &  7.1 & Gemma-3-27B     & 18.6 &  0.1 & 0.0 & 0.0 &  0.1 \\
Qwen3.6-35B-A3B & 38.1 & 15.6 & 3.7 & 0.8 & 11.2 & Qwen3-30B-A3B   & 16.0 & 11.9 & 6.4 & 0.3 &  4.8 \\
Qwen3.5-35B-A3B & 31.9 & 13.2 & 2.4 & 1.5 &  9.3 & Qwen3-32B       & 13.3 &  1.5 & 0.2 & 0.0 &  0.6 \\
\bottomrule
\end{tabular}
\end{table}

\subsection{Robustness and validity (extends Section~\ref{subsec:validity})}
\label{app:robustness_validity}

\paragraph{Cross-benchmark scores and correlations.}
\label{app:cross_benchmark}
The $16$-model overlap drops the three lowest-tier compute-efficient checkpoints (Gemma-3-27B, Qwen3-30B-A3B, Qwen3-32B) for which public scores are not available on most external benchmarks. Table~\ref{tab:cross_benchmark} reports per-model scores on seven external benchmarks spanning reasoning (GPQA-Diamond, HLE no-tools), code (SWE-Bench Pro/Public, Terminal-Bench~2.0, NL2Repo), and agent (BrowseComp, HLE with tools) categories. Each entry is the publicly-reported score for the specific checkpoint evaluated in Section~\ref{sec:experiments}; entries marked $^*$ are proxies imputed from a closely related sibling checkpoint where official scores for the exact variant are not available; $\dagger$ denotes a context-managed variant. Cells without verifiable public scores are marked ---. The Spearman rank correlations of each benchmark against AgentHop accuracy summarized in Section~\ref{subsec:validity} are GPQA-Diamond $\rho=+0.73$ ($n=15$), HLE no-tools $\rho=+0.72$ ($n=14$), Terminal-Bench~2.0 $\rho=+0.70$ ($n=15$), NL2Repo $\rho=+0.70$ ($n=10$), HLE with tools $\rho=+0.67$ ($n=10$).

\begin{table}[h]
\centering
\caption{Per-model scores on seven external benchmarks (publicly reported, normalised to percent) alongside AgentHop accuracy. Models grouped by tier and ranked by AgentHop accuracy. $^*$ denotes a proxy imputed from a closely related sibling checkpoint; $\dagger$ denotes a context-managed variant. \texttt{deepseek-chat} (the non-thinking V3.2 invocation) lacks public benchmark scores at the time of this submission and is reported as ---.}
\label{tab:cross_benchmark}
\scriptsize
\setlength{\tabcolsep}{4.5pt}
\renewcommand{\arraystretch}{0.95}
\begin{tabular}{l rrrrrrr r}
\toprule
Model & GPQA-D & HLE & SWE-B Pro & T-Bench~2.0 & NL2Repo & BrowseComp & HLE+tools & AgentHop \\
\midrule
\multicolumn{9}{l}{\textit{Closed Frontier}} \\
Gemini-3 Pro            & 91.9 & 37.5 & 43.3   & 56.9 & ---    & 59.2 & ---  & 89.1 \\
GPT-5.3 codex           & 92.6 & ---  & 56.8   & 77.3 & ---    & 77.3 & ---  & 82.5 \\
GPT-5.4                 & 92.8 & 39.8 & 57.7   & 75.1 & 41.3$^*$ & 82.7 & 52.1 & 79.0 \\
Claude Opus 4.6         & 91.3 & 40.0 & 57.3$^*$ & 65.4 & 49.8$^*$ & 84.0 & 53.1 & 77.5 \\
Claude Sonnet 4.6       & 89.9 & 33.2 & ---    & 59.1 & ---    & 74.0 & 49.0 & 76.6 \\
Gemini-3 Flash          & 90.4 & 33.7 & ---    & 47.6 & ---    & ---  & ---  & 54.7 \\
\midrule
\multicolumn{9}{l}{\textit{Open Frontier}} \\
GLM-5.1                 & 86.2 & 31.0 & 58.4   & 63.5 & 42.7   & 68.0 & 52.3 & 69.6 \\
DeepSeek-reasoner       & 79.9$^*$ & 19.8$^*$ & ---    & 37.7$^*$ & --- & 40.1$^*$ & 40.8$^\dagger$ & 59.5 \\
DeepSeek-chat           & ---  & ---  & ---    & ---  & ---    & ---  & ---  & 56.4 \\
Kimi-K2.5               & 87.6 & 30.1 & 50.7   & 50.8 & 32.0$^*$ & 60.6 & 50.2 & 44.1 \\
MiniMax-M2.7            & 87.0$^*$ & 28.0$^*$ & 56.2   & 57.0 & 39.8   & ---  & ---  & 41.3 \\
\midrule
\multicolumn{9}{l}{\textit{Compute-Efficient}} \\
Gemma-4-31B             & 84.3 & 19.5 & 35.7$^*$ & 42.9$^*$ & 15.5$^*$ & ---  & 26.5 & 64.8 \\
Qwen3.5-27B             & 85.5 & 24.3 & 51.2$^*$ & 41.6 & 27.3$^*$ & 61.0 & 48.5 & 48.2 \\
Qwen3.6-35B-A3B         & 86.0 & 21.4 & 49.5   & 51.5 & 29.4   & ---  & ---  & 38.1 \\
Qwen3.5-35B-A3B         & 84.2 & 22.4 & 44.6$^*$ & 40.5 & 20.5$^*$ & 61.0 & 47.4 & 31.9 \\
Gemma-4-26B-A4B         & 82.3 & 8.7  & 13.8$^*$ & 34.2$^*$ & 11.6$^*$ & ---  & 17.2 & 23.6 \\
\bottomrule
\end{tabular}
\end{table}

\paragraph{Run-to-run robustness on three multi-run checkpoints.}
\label{app:robustness}
For DeepSeek-chat, Gemma-4-31B, and GPT-5.3 codex we ran the full $1{,}011$-item evaluation three times under the same protocol with independent sampling. The agentic loop is multi-turn and stochastic by construction: each trajectory chains $8$--$16$ tool-call decisions before committing, and any single decision can re-route the rest of the trajectory. The natural expectation is therefore that headline accuracy will drift across runs. Empirically it does not. DeepSeek-chat and GPT-5.3 codex hold within $1$\,pp of accuracy across all three runs (std $\leq 0.009$), and the conversion-rate and per-trajectory-token figures are similarly tight. Gemma-4-31B is the exception: accuracy std $0.031$, conversion std $0.046$, and a $21.8$K-token spread per trajectory, a spread comparable to the cross-model gap between adjacent closed-frontier checkpoints. Table~\ref{tab:robustness} reports per-run values for accuracy and three representative axis metrics with across-run mean, standard deviation, and range; cross-model rank orderings on accuracy and on each diagnostic axis are preserved across all three runs.

\begin{table}[h]
\centering
\caption{Run-to-run figures across three runs for the three multi-run checkpoints. The two closed-frontier-style checkpoints (DeepSeek-chat, GPT-5.3 codex) show very low variance ($<0.011$ standard deviation on every metric except absolute tokens). Gemma-4-31B is more variable: accuracy std $0.032$, conversion std $0.047$, and notably tokens std $10.2$K (range $21.8$K), a spread comparable to the cross-model gap between adjacent closed-frontier checkpoints. Cross-model rank orderings on accuracy and on each of the four axes are nonetheless preserved across all three runs.}
\label{tab:robustness}
\small
\setlength{\tabcolsep}{5pt}
\begin{tabular}{l l rrr rrr}
\toprule
Model & Metric & Run 0 & Run 1 & Run 2 & Mean & Std & Range \\
\midrule
DeepSeek-chat
 & Accuracy        & 0.564 & 0.553 & 0.542 & 0.553 & 0.009 & 0.022 \\
 & Section recall  & 0.439 & 0.452 & 0.433 & 0.441 & 0.008 & 0.019 \\
 & Conversion      & 0.669 & 0.652 & 0.642 & 0.654 & 0.011 & 0.027 \\
 & Tokens (k)      & 156.7 & 156.1 & 155.1 & 156.0 & 0.7   & 1.6   \\
\midrule
Gemma-4-31B
 & Accuracy        & 0.648 & 0.585 & 0.581 & 0.605 & 0.031 & 0.067 \\
 & Section recall  & 0.275 & 0.242 & 0.251 & 0.256 & 0.014 & 0.033 \\
 & Conversion      & 0.795 & 0.722 & 0.685 & 0.734 & 0.046 & 0.110 \\
 & Tokens (k)      & 87.6  & 66.3  & 65.8  & 73.2  & 10.2  & 21.8  \\
\midrule
GPT-5.3 codex
 & Accuracy        & 0.825 & 0.818 & 0.810 & 0.818 & 0.006 & 0.015 \\
 & Section recall  & 0.380 & 0.371 & 0.370 & 0.374 & 0.004 & 0.010 \\
 & Conversion      & 0.888 & 0.888 & 0.888 & 0.888 & 0.000 & 0.000 \\
 & Tokens (k)      & 69.5  & 69.2  & 69.2  & 69.3  & 0.1   & 0.3   \\
\bottomrule
\end{tabular}
\end{table}

\subsection{Tool ablations (extends Section~\ref{subsec:ablations})}
\label{app:tool_ablations}

\paragraph{Parametric reliance: zero-tool baseline.}
\label{app:parametric}
To separate parametric capability from tool-augmented capability, we re-evaluate each model with all retrieval and reasoning tools removed: the model receives the question and four options and must commit to a letter from internal knowledge alone. Table~\ref{tab:zerotool_full} reports closed-book accuracy alongside full-tool accuracy and the resulting tool-conditioned lift ($\Delta$) for all 19 evaluated models, ranked by full-tool accuracy. Closed-book performance is itself a measured per-model property complementary to the headline accuracy, and the gap between closed-book and full-tool accuracy is what tools contribute over and above prior knowledge of the cited literature.

\textbf{Tool-conditioned lift scales inversely with closed-book strength.} Among models that produce valid answers without tools, weaker closed-book performers gain more from tool access, as Table~\ref{tab:zerotool_full} shows. DeepSeek-chat is the extreme: the largest absolute lift in the panel, ending within a few percentage points of Gemini-3 Pro's headline accuracy despite starting from a single-digit-teen closed-book floor. Closed-frontier checkpoints with stronger parametric coverage gain less because they are already answering more questions before retrieval ever helps. The pattern is consistent with the synthesis-axis interpretation in Section~\ref{subsec:main_results}: tools substitute for parametric coverage on items the model can synthesize from what it retrieves.

\textbf{Three checkpoints take a net loss from the agent loop.} Three panel members show negative tool-conditioned $\Delta$ in Table~\ref{tab:zerotool_full}: the seven-tool sandbox actively reduces their accuracy below the closed-book baseline. The mechanism differs in each case, but the diagnosis is the same: the agent loop disrupts a conversion path that closed-book inference handled directly. \emph{Gemini-3 Flash} crosses the 200K-token cap on nearly a third of trajectories before reaching \texttt{submit\_answer}, as detailed in Section~\ref{app:na_decomp_full}; \emph{Gemma-4-26B-A4B}'s sparse-active verbose generation drives the non-submission rate past forty percent; \emph{Qwen3-32B} falls into a metadata-only tool cycle that never reads paper content. None of these reflect a knowledge regression, parametric content is unchanged across conditions, but each documents a behavioral failure that only the full-tool sandbox surfaces.

\textbf{Below-chance closed-book commitments expose distractor susceptibility.} Ten of the nineteen checkpoints in Table~\ref{tab:zerotool_full} score below the random-four-option chance baseline of $25\%$ on closed-book, led by Kimi-K2.5 and the Qwen3 MoE variants. Their commits are systematically anti-correlated with the correct answer rather than random, which we attribute to vulnerability to the engineered distractors: with no read access to the supplied papers, the \texttt{no\_context} distractor, designed to be parametrically plausible, is confidently picked over the gold. The implication for panel-level reading is that a low closed-book number must be paired with the section-recall column of Table~\ref{tab:main_results} to be interpretable. It can reflect either weak parametric coverage (which tools repair) or full-tool behavioral failure (which they do not), and only the joint reading of the two columns disambiguates the two. Because a near-chance checkpoint cannot express a gradient along any stratification axis, closed-book comparisons in this paper are read per tier.

\begin{table}[h]
\centering
\caption{Parametric (zero-tool) vs.\ full-tool accuracy across the 19-model panel, tier-grouped and ranked by full-tool accuracy within each tier. Chance baseline is $25\%$; negative $\Delta$ indicates that the seven-tool sandbox reduces accuracy below the closed-book baseline.}
\label{tab:zerotool_full}
\scriptsize
\setlength{\tabcolsep}{4pt}
\renewcommand{\arraystretch}{0.95}
\begin{tabular}{lrrr @{\hskip 1.5em} lrrr}
\toprule
Model & Zero & Full & $\Delta$ & Model & Zero & Full & $\Delta$ \\
\midrule
\multicolumn{4}{l}{\textit{Closed Frontier}}                & \multicolumn{4}{l}{\textit{Open Frontier}} \\
Gemini-3 Pro      & 73.4 & 89.1 & $+15.7$ & GLM-5.1            & 37.6 & 69.6 & $+32.0$ \\
GPT-5.3 codex     & 48.6 & 82.5 & $+33.9$ & DeepSeek-reasoner  & 25.0 & 59.5 & $+34.5$ \\
GPT-5.4           & 59.3 & 79.0 & $+19.7$ & DeepSeek-chat      & 18.3 & 56.4 & $+38.1$ \\
Claude Opus 4.6   & 55.2 & 77.5 & $+22.4$ & Kimi-K2.5          & 12.1 & 44.1 & $+32.0$ \\
Claude Sonnet 4.6 & 48.5 & 76.6 & $+28.1$ & MiniMax-M2.7       & 14.5 & 41.3 & $+26.8$ \\
Gemini-3 Flash    & 57.3 & 54.7 & $-2.6$  &                    &      &      &         \\
\midrule
\multicolumn{8}{l}{\textit{Compute-Efficient}} \\
Gemma-4-31B     & 31.7 & 64.8 & $+33.1$ & Gemma-4-26B-A4B & 24.5 & 23.6 & $-0.9$  \\
Qwen3.5-27B     & 17.1 & 48.2 & $+31.1$ & Gemma-3-27B     &  7.8 & 18.6 & $+10.8$ \\
Qwen3.6-35B-A3B & 14.0 & 38.1 & $+24.0$ & Qwen3-30B-A3B   & 13.4 & 16.0 & $+2.7$  \\
Qwen3.5-35B-A3B & 13.2 & 31.9 & $+18.8$ & Qwen3-32B       & 18.9 & 13.3 & $-5.6$  \\
\bottomrule
\end{tabular}
\end{table}

\paragraph{Zero-tool accuracy by exposure strata.}
\label{app:zt_stratification}
Table~\ref{tab:zt_strata} stratifies closed-book accuracy by gold-paper citation count, seed publication year, and venue. Citation counts are the construction-time values recorded in the pipeline, and each sample takes the maximum over its gold papers. No axis shows a monotone gradient for the closed-frontier tier, and the most cited quartile is the lowest. The open and compute-efficient tiers sit near the $25$\% chance floor, so their strata are reported for completeness rather than comparison.

\begin{table}[h]
\centering
\caption{Zero-tool accuracy (\%) by stratum, with per-tier decomposition. Quartiles are
rank-based over the $1{,}011$ samples; ranges give citation counts.}
\label{tab:zt_strata}
\small
\begin{tabular}{llrrrr}
\toprule
Axis & Stratum & $n$ & Closed & Open & Comp.-eff. \\
\midrule
Citations & Q1 (1--224)        & 252 & 57.5 & 21.7 & 19.2 \\
          & Q2 (226--585)      & 253 & 59.2 & 22.2 & 17.8 \\
          & Q3 (586--1{,}554)  & 253 & 57.1 & 19.9 & 16.5 \\
          & Q4 (1{,}554--54{,}992) & 253 & 54.4 & 22.2 & 16.7 \\
\midrule
Year      & 2022 &  72 & 53.5 & 18.3 & 19.6 \\
          & 2023 & 355 & 54.7 & 19.7 & 15.9 \\
          & 2024 & 544 & 58.9 & 23.4 & 18.5 \\
          & 2025 &  40 & 58.8 & 17.5 & 16.6 \\
\midrule
Venue     & ACL     & 113 & 57.5 & 21.2 & 18.0 \\
          & CVPR    & 133 & 55.8 & 19.1 & 16.1 \\
          & ECCV    & 102 & 57.4 & 21.8 & 16.3 \\
          & EMNLP   & 102 & 57.5 & 19.2 & 20.5 \\
          & ICLR    & 198 & 58.0 & 23.1 & 17.9 \\
          & ICML    &  96 & 55.2 & 22.1 & 15.8 \\
          & NAACL   &  79 & 61.4 & 25.3 & 22.0 \\
          & NeurIPS & 153 & 55.7 & 22.0 & 16.7 \\
          & SIGIR   &  35 & 53.8 & 16.0 & 13.6 \\
\bottomrule
\end{tabular}
\end{table}

\paragraph{Three-checkpoint ablation protocol.}
\label{app:ablations}
For the three checkpoints in Section~\ref{subsec:ablations} (DeepSeek-chat, Gemma-4-31B, GPT-5.3 codex), the \emph{no-search} variant removes only \texttt{search\_papers} from the seven-tool sandbox, retaining the other four literature-interacting tools (\texttt{get\_paper\_info}, \texttt{get\_references}, \texttt{list\_sections}, \texttt{read\_section}) plus \texttt{think} and \texttt{submit\_answer}; the agent can still navigate by following references but cannot retrieve paper IDs by keyword. The \emph{zero-tool} variant removes the sandbox entirely (system prompts in Appendix~\ref{app:prompts}, Prompts 8 \& 9). The cost-bounded protocol prevents running these ablations on the full $19$-model panel; the three-checkpoint cohort is chosen to span the accuracy range, and we do not extrapolate to the broader pool. Per-checkpoint deltas are reported in Table~\ref{tab:ablation_main} of Section~\ref{subsec:ablations}.

%% file: sections/Apdx_F.tex

\section{Prompts and Tool Catalogue}
\label{app:prompts}

This appendix collects the verbatim prompts used in the construction pipeline and at evaluation time, alongside the tool catalogue exposed to the agent at evaluation time. Dynamic budget templates and per-provider formatting wrappers are omitted for brevity but available in the released harness.

\subsection{Tool catalogue}
\label{app:tools}

The agent has access to seven tools at every step. Each tool is presented to the model as a function-calling schema with the input arguments listed in Table~\ref{tab:tools}; the per-call cost is deducted from the 30-point budget at invocation. Cost asymmetry is intentional: one-point navigation tools encourage broad exploration, the five-point reading tool forces selective use, and the two free tools (\texttt{think}, \texttt{submit\_answer}) are unmetered so neither reasoning nor answering can be priced out.

\begin{table}[h]
\centering
\caption{The seven tools exposed to the agent. Costs are per call, deducted from the per-trajectory 30-point budget at invocation. Identical schemas, identical input typings, and identical natural-language descriptions are presented to every model.}

\small
\setlength{\tabcolsep}{4pt}
\renewcommand{\arraystretch}{1.15}
\begin{tabular}{l c p{0.30\textwidth} p{0.40\textwidth}}
\toprule
\textbf{Tool} & \textbf{Cost} & \textbf{Input} & \textbf{Returns} \\
\midrule
\texttt{get\_paper\_info} & 1 & \texttt{paper\_id}                                            & Title, abstract, year, and venue of the paper. \\
\texttt{get\_references}  & 1 & \texttt{paper\_id}                                            & List of cited papers with their IDs, titles, years, and the citation context (the sentence in which they were cited). \\
\texttt{list\_sections}   & 1 & \texttt{paper\_id}                                            & Section headers with alphabet aliases and per-section character counts, for example \textsf{(A) Introduction (5869 chars)}. Aliases may be passed to \texttt{read\_section} in place of a full header, and the counts let an agent price a 5-point read before paying for it. \\
\texttt{search\_papers}  & 1 & \texttt{query} (str), \texttt{top\_k} ($\leq 10$, default 5)  & Top-$k$ matching papers from the available pool, each with title and abstract. \\
\texttt{read\_section}    & 5 & \texttt{paper\_id}, \texttt{section}                          & Full text of the requested section. The \texttt{section} argument accepts the full header, the alphabet alias from \texttt{list\_sections}, or a unique case-insensitive substring. \\
\texttt{think}            & 0 & \texttt{thought} (str)                                        & Acknowledgement; the thought is appended to the trajectory transcript and remains available for the agent's later reasoning steps. \\
\texttt{submit\_answer}   & 0 & \texttt{option} (\texttt{A}/\texttt{B}/\texttt{C}/\texttt{D}) & Terminates the trajectory and records the agent's final choice. \\
\bottomrule
\end{tabular}
\label{tab:tools}
\end{table}

\subsection{Error handling and budget enforcement}
\label{app:error_handling}

\paragraph{Budget rejection.}
Each tool call deducts its per-call cost from the trajectory's $30$-point pool at invocation. If the remaining budget is insufficient to cover a call's cost, the harness rejects the call with the message: ``\textsf{REJECTED: Insufficient budget (\textit{X} remaining, \textit{tool} costs \textit{Y}). You cannot use this tool. Call \texttt{submit\_answer} immediately with your best guess based on evidence gathered so far.}'' No cost is deducted on rejection. The agent sees the rejection in its conversation history and may issue \texttt{submit\_answer} on the next turn. If a second budget-rejected tool call occurs immediately after the first, the trajectory terminates as \texttt{terminated\_by\,=\,"budget"}; the first successful (non-rejected) call resets this counter to zero. When budget reaches zero before the agent's next turn, the harness additionally appends ``\textsf{Budget exhausted. You MUST call \texttt{submit\_answer} now with your best guess}'' to the next user-side input as a soft nudge. The agent is informed of the total budget in the system prompt, and every tool result carries a one-line status footer with the current turn, cumulative token usage, and budget spent (e.g., \textsf{Turn: 8 | Tokens used: 23,379 | Budget used: 20}); resource-management behavior is therefore measured under full disclosure of resource state.

\paragraph{Token-limit enforcement.}
Cumulative tokens are checked against the $200$K cap after every assistant response. If the cap is exceeded, the trajectory terminates immediately with \texttt{terminated\_by\,=\,"token\_limit"}.

\paragraph{Max-turns enforcement.}
The trajectory terminates with \texttt{terminated\_by\,=\,"max\_turns"} once the agent has issued $20$ assistant turns without calling \texttt{submit\_answer}. The live turn count appears in the same per-result status footer.

\paragraph{Termination categories.}
Trajectories end via one of five outcomes: \texttt{"answer"} (the agent calls \texttt{submit\_answer}), \texttt{"budget"} (two consecutive budget rejections), \texttt{"token\_limit"} ($200$K cap hit), \texttt{"max\_turns"} ($20$-turn cap hit), or \texttt{"error"} (five consecutive non-transient API errors). The first four are expected outcomes and constitute the four submission/non-submission categories analyzed throughout the paper; \texttt{"error"} is reserved for systematic-failure review and is rare in practice.

\paragraph{Sample-level retry on hard error.}
A sample is permitted up to two retries (three attempts total) on hard errors: API authentication failure, persistent provider 5xx, or five consecutive non-transient API errors within a single attempt. Transient errors (rate-limit, temporary 5xx) trigger an in-place retry without consuming the turn budget or a sample-retry counter. Tool-level malformations (unknown tool name, missing required argument, malformed JSON envelope) are surfaced to the agent as an error message in the tool result and do not retry the sample.

The remaining subsections list the verbatim system prompts used at every stage of the pipeline and at evaluation time. Each prompt is shown inside a breakable framed box with a caption identifying the pipeline stage at which it is invoked. Prompts appear in pipeline order: chain-screening judge (Stage 2), question generation (single-target then multi-target), MCQ distractor generation (paper-grounded then parametric), the two automated filters (shortcut detection then consensus), the GPT-5.4 audit-prep generator that drives the human audit, and finally the agent and ablation system prompts used at evaluation time. The budget block is static across samples and appears in full in the agent system prompt.

\begin{tcolorbox}[
  breakable,
  enhanced jigsaw,
  colback=gray!5,
  colframe=gray!50,
  fonttitle=\bfseries,
  before upper={\small\ttfamily}
]
\captionof{table}{Stage 2 chain-screening prompt. GPT-5.4 scores each candidate (seed, hop1, candidate) chain on a 1--5 ``curiosity gap'' scale conditioned on the abstracts and citation path. Only chains scoring $\geq 4$ are retained as MCQ-generation candidates.}

\begin{verbatim}
You are screening candidate papers for a multi-hop research QA benchmark.
Given a seed paper and a candidate paper reachable through citation links,
evaluate whether the candidate is a good question target.

A good question target has:
1. A specific curiosity gap: a reader of the seed would have a natural reason
   to seek this paper
2. Specific factual content: the candidate's abstract mentions concrete
   results, numbers, metrics, or findings
3. Non-obviousness: the answer requires actually navigating to and reading
   the candidate paper

You will receive the seed paper's abstract, the candidate paper's abstract,
and the citation path connecting them. Focus on the content of the abstracts
and the research narrative implied by the path.

=== FEW-SHOT EXAMPLES ===

Example 1 - KEEP (score: 4)
Seed: "RAG with Knowledge Graphs for Customer Service QA" (2024)
Candidate: "UniKGQA: Unified Retrieval and Reasoning for Multi-hop KGQA" (2022)
Path: Seed -> "Reasoning on Graphs" -> Candidate
Evaluation: {"score": 4, "curiosity_gap": "Seed uses KG-augmented RAG; hop1
  discusses KG reasoning for LLMs and explicitly states it improves on prior
  KGQA methods; candidate is the SOTA method being improved upon - a reader
  would want to know what the baseline approach was", "candidate_fact":
  "UniKGQA unifies retrieval and reasoning in both architecture and parameter
  learning for multi-hop KGQA", "reasoning": "The citation path shows a clear
  improvement relationship with specific technical claims. The candidate has
  concrete contributions that a seed reader would naturally want to understand."}

Example 2 - REJECT (score: 2)
Seed: "MIRAGE: Metric-Intensive Benchmark for RAG Evaluation" (2025)
Candidate: "How Much Knowledge Can You Pack into Parameters of a Language
  Model?" (2020)
Path: Seed -> "RAG for Knowledge-Intensive NLP Tasks" -> Candidate
Evaluation: {"score": 2, "curiosity_gap": "The connection is tangential -
  hop1 cites the candidate as a closed-book QA baseline for comparison, not
  as a follow-up research question", "candidate_fact": "Mentions that storing
  knowledge in parameters scales with model size, but no specific metrics in
  abstract", "reasoning": "The candidate is cited as a contrasting paradigm
  (retrieval vs parametric), not as part of a research trajectory a seed
  reader would naturally follow."}

Example 3 - BORDERLINE (score: 3)
Seed: "MIRAGE: Metric-Intensive Benchmark for RAG Evaluation" (2025)
Candidate: "Dense Passage Retrieval for Open-Domain Question Answering" (2020)
Path: Seed -> "RAG for Knowledge-Intensive NLP Tasks" -> Candidate
Evaluation: {"score": 3, "curiosity_gap": "RAG uses DPR as its retriever -
  a reader might want to understand the retrieval component, but this is more
  of a technical dependency than a research curiosity gap", "candidate_fact":
  "Dense retriever outperforms BM25 by 9-19% absolute in top-20 passage
  retrieval accuracy", "reasoning": "The candidate has specific metrics but
  the connection from the seed is about implementation detail (which
  retriever RAG uses) rather than a research question. Useful but generic."}

=== END EXAMPLES ===

Score 1-5:
1 = no viable question (group citation, no specific facts, or answer is
    obvious from seed)
2 = weak (vague connection or abstract lacks specific findings)
3 = possible but generic (connection exists but question would be shallow)
4 = good (specific curiosity gap + concrete answerable facts in candidate)
5 = excellent (clear research narrative + specific metrics/findings)

Respond as JSON only:
{
  "score": N,
  "curiosity_gap": "what specific question would a seed reader have about
                    this paper?",
  "candidate_fact": "a specific fact/number from the candidate's abstract,
                     or null",
  "reasoning": "one sentence explaining your score based on the abstracts
                and path"
}
\end{verbatim}
\end{tcolorbox}
\label{prompt:chain_judge}

\begin{tcolorbox}[
  breakable,
  enhanced jigsaw,
  colback=gray!5,
  colframe=gray!50,
  fonttitle=\bfseries,
  before upper={\small\ttfamily}
]
\captionof{table}{System prompt used at the question-generation stage of the construction pipeline to generate single-target (depth-1 and depth-2) questions whose answer lies in a single terminal paper reachable from the seed.}

\begin{verbatim}
You are creating research questions that model how a researcher
THINKS while reading a paper.

A researcher reads the seed paper, notices a specific claim,
comparison, or referenced technique, and follows citations to the
terminal paper where they find the answer.

Your job: given a citation chain (seed -> ... -> terminal) with full
paper content, create a question that captures this natural research
curiosity. Each prompt includes a QUESTION MODE specifying the
citation relationship and cognitive skill to target.

=== STEP 1: UNDERSTAND THE SEED PAPER ===

Before generating any question, first read the seed paper carefully
and identify:
1. What is the paper's core contribution or thesis?
2. What specific claims, comparisons, or design choices does it make?
3. How does it reference or rely on the cited works in this chain?

Articulate this understanding in the "seed_understanding" field. This
grounds everything that follows -- a good question cannot come from a
superficial reading.

=== ANSWER RULES ===

The answer must be 1-2 sentences stating a specific finding, design
choice, or result as a direct factual claim. Write as if asserting
the fact itself -- do NOT attribute it to a paper ("the paper
reports...", "the cited work shows...", "the authors found..."). Just
state the finding.

- WRONG: "The cited paper reports that the model achieves 94.1%
  accuracy on COCO."
- WRONG: "According to the referenced work, tiling reduces memory
  by 3x."
- RIGHT: "The model achieves 94.1% accuracy on COCO with a ResNet-50
  backbone."
- RIGHT: "Tiling the attention computation reduces peak memory usage
  by 3x."

Do not give a bare number; describe the finding. Do not speculate or
add reasoning beyond what the terminal paper states.

If no good answer exists in the terminal paper content, output
"answer": null.

=== ANSWER_CONTEXT RULES ===

The answer_context field must be EXACT TEXT copied from the terminal
paper content above. Do not add any framing, attribution, or
modification:
- WRONG: "From Table 3: the model achieves 94.1%..."
- WRONG: "From the terminal paper: 'We use T=1000...'"
- WRONG: "The authors report that..."
- RIGHT: "We use T = 1000 for all experiments. We set the forward
  process variances to constants increasing linearly from
  beta_1 = 10^-4 to beta_T = 0.02."

If the answer comes from a table, copy the relevant row(s) exactly
as they appear in the table content. Do not paraphrase or summarize
table data.

=== QUESTION RULES ===

The question must be GROUNDED in a specific detail from the seed
paper -- a claim, comparison, design choice, limitation, or cited
result.

CRITICAL -- SEED DEPENDENCY: The question must require information
from the seed paper to answer, not just as a navigation clue but as
CONTEXT that determines what to look for in the terminal. The
question text should reference a concrete detail from the seed (a
number, a claim, a comparison, a stated limitation) that CONSTRAINS
which part of the terminal paper is relevant.

Self-check: if someone receives ONLY the terminal paper and this
question (without the seed), would they immediately know which of
the terminal's many findings to focus on? If yes, the seed is just a
pointer -- rewrite so the seed's specific claim narrows the answer.

Good pattern: "Seed claims X about cited work -> what does cited
work report about X?"
Bad pattern: "Seed mentions cited work -> what does cited work do?"

Naming:
- You MAY freely name the seed paper, its methods/models, and
  intermediate papers
- MUST NOT name the terminal paper by title -- the agent must
  discover it through navigation

Phrasing -- do NOT use any of these patterns:
- "In [paper], the authors..."
- "While reading [paper], a researcher..."
- "A researcher reading [paper]..."
- "To contextualize..."

Instead, write direct questions grounded in specific paper content.

Output as JSON:
{
  "seed_understanding": "2-3 sentences: the seed paper's core
    contribution and how the citation chain relates to it",
  "seed_detail": "The specific claim/comparison/detail from the seed
    paper that motivates the question",
  "seed_dependency": "Explain how the seed's detail constrains what
    to look for in the terminal -- why can't someone answer this
    from the terminal alone?",
  "question": "The research question (2-4 sentences, direct
    phrasing)",
  "answer": "1-2 sentence answer drawn from the terminal paper",
  "answer_context": "The sentence/paragraph containing the answer
    (copy verbatim from terminal content)",
  "hops_required": "what each hop contributes to answering"
}
\end{verbatim}
\end{tcolorbox}
\label{prompt:bfs_gen}

\medskip

\begin{tcolorbox}[
  breakable,
  enhanced jigsaw,
  colback=gray!5,
  colframe=gray!50,
  fonttitle=\bfseries,
  before upper={\small\ttfamily}
]
\captionof{table}{System prompt used at the question-generation stage of the construction pipeline to generate multi-target questions whose answer requires synthesizing claims across two or three cited papers.}

\begin{verbatim}
You are creating research questions that require SYNTHESIZING
information across multiple cited papers.

A researcher reads the seed paper, notices a passage discussing or
comparing two or three cited works, and realizes they need to read
BOTH cited works and combine their findings to fully understand the
comparison.

Your job: given a seed paper and 2-3 target papers it cites (with
full content), create a question that:
1. Guides the reader from the seed paper TOWARD the target papers
   through narrative clues
2. REQUIRES reading multiple target papers to answer

Each prompt includes a QUESTION MODE specifying the citation
relationship and cognitive skill to target.

=== STEP 1: UNDERSTAND THE SEED PAPER ===

Before generating any question, first read the seed paper carefully
and identify:
1. What is the paper's core contribution or thesis?
2. What specific passage discusses or compares the target papers
   together?
3. What role does each target paper play in the seed's narrative
   (baseline, inspiration, competing approach, etc.)?

Articulate this understanding in the "seed_understanding" field.
This grounds everything that follows -- a good synthesis question
cannot come from a superficial reading.

=== ANSWER RULES ===

The answer must be 1-2 sentences combining specific observations
from both papers as direct factual claims. Write as if asserting the
facts -- do NOT attribute them to papers ("the paper reports...",
"the cited work shows...", "the authors found..."). Just state the
findings.

- WRONG: "The first paper reports 94.1% AP while the second paper
  achieves 91.3% AP."
- RIGHT: "The sparse detector achieves 94.1% AP while the
  conditional query approach reaches 91.3% AP on COCO val2017."

MUST NOT name target papers -- use role descriptions ("the first
approach", "the sparse detector"). Do not speculate or add reasoning
beyond what the papers state.

If no good synthesis question exists, output null for all fields.

=== ANSWER_SOURCES RULES ===

Each "fact" in answer_sources must be EXACT TEXT copied from that
paper's content above. Do not add any framing, attribution, or
modification:
- WRONG: "Table 3 shows that the model achieves 94.1%..."
- WRONG: "The authors report that..."
- RIGHT: "achieves 94.1% accuracy on COCO val2017 with a ResNet-50
  backbone"

If the fact comes from a table, copy the relevant row(s) exactly as
they appear in the table content. Do not paraphrase or summarize
table data.

=== QUESTION RULES ===

The question must be grounded in a specific passage from the seed
paper where multiple cited works are discussed together. Describe
each target paper by the ROLE or CHARACTERIZATION the seed gives it
-- not by name.

CRITICAL -- SEED DEPENDENCY: The question must require information
from the seed paper to answer, not just as a way to identify which
papers to read. The seed should provide context that CONSTRAINS what
to look for across the target papers -- a specific comparison frame,
a claim to verify, or a characterization that narrows the relevant
findings.

Self-check: if someone receives ONLY the target papers and this
question (without the seed), would they immediately know which
findings to compare and how? If yes, the seed is unnecessary --
rewrite so the seed's specific framing determines the answer.

Naming:
- You MAY freely name the SEED paper, its methods, models, and
  datasets
- MUST NOT name target papers by title, method name, model name, or
  acronym
- Instead, DESCRIBE each target paper by the role the seed gives it:
    - "a sparse detector that learns proposals directly" (not
      "Sparse R-CNN")
    - "an approach that accelerates convergence via conditional
      spatial queries" (not "Conditional DETR")
- The description must be specific enough that a reader of the seed
  paper can identify which citation to follow

Phrasing -- do NOT use any of these patterns:
- "In [paper], the authors..."
- "While reading [paper], a researcher..."
- "A researcher reading [paper]..."
- "To contextualize..."

Instead, write direct questions grounded in specific paper content.

Output as JSON:
{
  "seed_understanding": "2-3 sentences: the seed paper's core
    contribution and how the target papers relate to it",
  "seed_detail": "The specific passage from the seed paper
    motivating this question",
  "seed_dependency": "Explain how the seed's framing constrains what
    to look for across target papers -- why can't someone answer
    this from the targets alone?",
  "question": "The synthesis question (2-4 sentences, target papers
    described by role not name)",
  "answer": "1-2 sentence answer combining observations from both
    papers (no target paper names)",
  "answer_sources": {
    "paper_1": {"arxivId": "...", "fact": "specific fact from
      paper 1"},
    "paper_2": {"arxivId": "...", "fact": "specific fact from
      paper 2"}
  },
  "why_multi_paper": "Brief explanation of why reading multiple
    papers is required"
}
\end{verbatim}
\end{tcolorbox}
\label{prompt:dfs_gen}

\medskip

\begin{tcolorbox}[
  breakable,
  enhanced jigsaw,
  colback=gray!5,
  colframe=gray!50,
  fonttitle=\bfseries,
  before upper={\small\ttfamily}
]
\captionof{table}{Paper-grounded distractor system prompt used at the MCQ-generation stage of the construction pipeline. Invoked twice per chain: once with the seed paper text to produce the \texttt{seed\_only} distractor (a finding genuinely supported by the seed but irrelevant to the cited claim), and once with a sibling target paper's text to produce the \texttt{wrong\_paper} distractor (a real claim from a related-but-incorrect paper).}

\begin{verbatim}
You answer research questions using the provided paper content.

=== ANSWER RULES ===

The answer must be 1-2 sentences describing a specific finding,
design choice, or result from the paper. It must be drawn directly
from what the paper states -- do not speculate, interpret, or add
reasoning beyond what the paper says. Do not give a bare number;
describe the finding.

If the provided content does not directly address the question, use
the most relevant information available to construct a specific,
affirmative answer. NEVER say "the paper does not", "does not
discuss", "does not specify", "does not mention", "not directly
addressed", or any similar qualification. Always answer with what IS
in the content, never with what is missing.

=== ANSWER_CONTEXT RULES ===

The answer_context field must be EXACT TEXT copied from the paper
content above. Do not add any framing, attribution, or modification:
- WRONG: "From Table 3: the model achieves 94.1%..."
- WRONG: "The authors report that..."
- RIGHT: "We use T = 1000 for all experiments. We set the forward
  process variances to constants increasing linearly from
  beta_1 = 10^-4 to beta_T = 0.02."

If the answer comes from a table, copy the relevant row(s) exactly
as they appear in the table content. Do not paraphrase or summarize
table data.

Output as JSON:
{"answer": "your 1-2 sentence answer", "answer_context": "exact text from source"}
\end{verbatim}
\end{tcolorbox}
\label{prompt:mcq_grounded}

\medskip

\begin{tcolorbox}[
  breakable,
  enhanced jigsaw,
  colback=gray!5,
  colframe=gray!50,
  fonttitle=\bfseries,
  before upper={\small\ttfamily}
]
\captionof{table}{Parametric (closed-book) distractor system prompt used at the MCQ-generation stage of the construction pipeline to produce the \texttt{no\_context} distractor: a confident, surface-plausible answer drawn solely from the model's parametric knowledge, with no paper text in the prompt.}

\begin{verbatim}
You answer research questions based on your knowledge of the
literature.

=== ANSWER RULES ===

The answer must be 1-2 sentences describing a specific finding,
design choice, or result. Be concrete and specific -- include
numbers, method names, and dataset names where possible. Do not give
a bare number; describe the finding.

NEVER say "I don't know", "I'm not sure", or any similar
qualification. Always give a specific, confident answer.

Output as JSON:
{"answer": "your 1-2 sentence answer"}
\end{verbatim}
\end{tcolorbox}
\label{prompt:mcq_parametric}

\medskip

\begin{tcolorbox}[
  breakable,
  enhanced jigsaw,
  colback=gray!5,
  colframe=gray!50,
  fonttitle=\bfseries,
  before upper={\small\ttfamily}
]
\captionof{table}{Shortcut-detection-filter system prompt used at the model-ensemble filtering stage. The 3-model ensemble (GPT-4.1, Claude Sonnet 4.6, \texttt{deepseek-chat}) is asked to answer the MCQ given access to only one paper at a time: the seed paper or the terminal target paper. Samples answerable from a single paper alone are discarded.}

\begin{verbatim}
You are answering a multiple-choice question about academic
research. You have access to ONE paper. Use ONLY the provided paper
to answer. If no option fits well, pick the closest one.

You MUST respond with ONLY this JSON and nothing else:
{"answer": "X"} where X is A, B, C, or D.
\end{verbatim}
\end{tcolorbox}
\label{prompt:filter_ab}

\medskip

\begin{tcolorbox}[
  breakable,
  enhanced jigsaw,
  colback=gray!5,
  colframe=gray!50,
  fonttitle=\bfseries,
  before upper={\small\ttfamily}
]
\captionof{table}{Consensus-filter system prompt used at the model-ensemble filtering stage. The 3-model ensemble is asked the MCQ \emph{without} any paper context. Samples on which the ensemble fails to converge on the gold answer are flagged for closer review and feed into the consensus-tier assignment described in Appendix~\ref{app:pipeline}.}

\begin{verbatim}
You are answering a multiple-choice question about academic
research. You have access to all papers needed to answer. Read
carefully and select the correct answer.

You MUST respond with ONLY this JSON and nothing else:
{"answer": "X"} where X is A, B, C, or D.
\end{verbatim}
\end{tcolorbox}
\label{prompt:filter_c}

\medskip

\begin{tcolorbox}[
  breakable,
  enhanced jigsaw,
  colback=gray!5,
  colframe=gray!50,
  fonttitle=\bfseries,
  before upper={\small\ttfamily}
]
\captionof{table}{GPT-5.4 audit-prep system prompt used to generate the structured guide that orients human auditors. Run once per post-filter sample with the full seed and target paper(s) attached, the prompt produces a JSON record (claim decomposition, chain or synthesis verification, and section-level recall labels) that is presented to human auditors during the Gradio audit pass and stored as the released ground truth for section-recall metrics.}

\begin{verbatim}
You are an audit assistant for AgentHop, a multi-hop scientific
reasoning benchmark.

You will receive:
- A benchmark question designed to test citation-chain navigation
- The correct answer to that question
- Full content of the seed paper and target paper(s), including
  appendix sections

Your job is to verify the QUERY and CORRECT ANSWER only. You are NOT
selecting an answer from multiple options.

The sample type will be indicated as "single-target" or
"multi-target". Apply the type-specific checks described below.

Return a JSON object with the following structure:

{
  "query_quality": {
    "verdict": "pass" | "warn" | "fail",
    "natural": true | false,
    "unambiguous": true | false,
    "requires_target": true | false,
    "note": "Brief explanation of any issues with the question
      wording, clarity, or scope. Empty string if none."
  },
  "answer_validity": {
    "verdict": "pass" | "warn" | "fail",
    "claims": [
      {
        "claim": "A specific factual assertion from the correct
          answer",
        "source_paper": "Which paper contains this claim (seed
          title or target title)",
        "source_section": "Exact section header where this claim
          is found",
        "verbatim_match": true | false,
        "verified": true | false,
        "note": "Discrepancy detail if not verified, or empty
          string"
      }
    ]
  },
  "chain_coherence": {
    "verdict": "pass" | "warn" | "fail",
    "seed_to_bridge": "Does the seed paper's citation context
      naturally point toward the bridge paper? Explain briefly.",
    "bridge_to_terminal": "Does the bridge paper lead to the
      terminal? Is the bridge necessary or could you skip it?",
    "bridge_necessary": true | false,
    "note": "Any issues with the citation chain logic. Empty
      string if none."
  },
  "synthesis_check": {
    "verdict": "pass" | "warn" | "fail",
    "both_targets_needed": true | false,
    "target_1_contribution": "What fact/evidence does target 1
      contribute to the answer?",
    "target_2_contribution": "What fact/evidence does target 2
      contribute to the answer?",
    "genuine_synthesis": true | false,
    "note": "Is this genuine cross-paper synthesis or just two
      independent facts? Empty string if fine."
  },
  "section_recall_labels": [
    {
      "paper": "Paper title (as provided in the content headers)",
      "arxiv_id": "arxiv ID if identifiable from the content
        header, else empty string",
      "section": "Section header containing answer-relevant
        evidence",
      "relevance": "direct" | "supporting",
      "what_it_contains": "Brief description of what evidence this
        section provides for the answer"
    }
  ],
  "structural_integrity": {
    "verdict": "pass" | "warn" | "fail",
    "missing_sections": ["Section names that appear to be missing
      from the provided content, if any"],
    "note": "Brief explanation of structural issues, or empty
      string"
  }
}

Type-specific instructions:

FOR SINGLE-TARGET SAMPLES:
- Focus on "chain_coherence". The key question is whether the
  seed -> bridge -> terminal path is natural and necessary.
- Check: does the seed paper's text contain a citation or reference
  that would lead a reader toward the bridge paper?
- Check: does the bridge paper connect to the terminal, or could you
  reach the terminal directly from the seed?
- Check: is the answer found ONLY in the terminal paper, not in the
  seed or bridge?
- Set "synthesis_check" fields to null -- not applicable for
  single-target.

FOR MULTI-TARGET SAMPLES:
- Focus on "synthesis_check". The key question is whether BOTH
  target papers are genuinely needed.
- Check: does the answer combine information from both targets, or
  could one target alone suffice?
- Check: is the synthesis genuine (comparing, contrasting,
  combining) or superficial (two unrelated facts)?
- Set "chain_coherence" fields to null -- not applicable for
  multi-target.

General guidelines:
- For "claims": decompose the correct answer into individual factual
  assertions. Each number, comparison, method name, or result is a
  separate claim.
- For "section_recall_labels": list ALL sections across ALL provided
  papers that contain evidence relevant to the answer. "direct" =
  the claim is stated here; "supporting" = provides context needed
  to interpret the claim.
- For "requires_target": true if the question cannot be answered
  from the seed paper alone (this should almost always be true).
- For "verbatim_match": true if the claim text appears nearly
  word-for-word in the source section.
- Be precise with section names -- use the exact header as it
  appears in the provided content.

IMPORTANT: Return ONLY the JSON object. No markdown, no explanation
outside the JSON.
\end{verbatim}
\end{tcolorbox}
\label{prompt:audit}

\medskip

\begin{tcolorbox}[
  breakable,
  enhanced jigsaw,
  colback=gray!5,
  colframe=gray!50,
  fonttitle=\bfseries,
  before upper={\small\ttfamily}
]
\captionof{table}{Agent system prompt used at evaluation time in the full-tool condition. Identical text is presented to every model; the placeholders \texttt{\{tools\_block\}} and \texttt{\{task\_block\}} are filled at request time with the seven-tool catalogue of Table~\ref{tab:tools} and the per-sample question with four options. The budget block is identical for every sample and is printed in full above.}

\begin{verbatim}
You are a research assistant tasked with answering a multi-hop
scientific question. You will navigate a citation graph of academic
papers using a set of tools. Your goal is to gather sufficient
evidence from the papers to select the correct answer.

## Task Overview
You are given a seed paper as your starting point. The answer
requires information from one or more papers reachable through the
citation graph -- papers cited by the seed, or papers cited by those
papers (up to 2 hops). You must explore the graph strategically,
read relevant sections, and synthesize evidence to choose the best
answer.

{tools_block}

## Constraints
You operate under three independent limits. The run terminates
when any one is hit:
- **Budget**: 30 points total. Navigation tools cost 1 pt;
  read_section costs 5 pts.
- **Turns**: at most 20 conversational turns.
- **Tokens**: at most 200,000 cumulative prompt + completion
  tokens per sample.

After every tool result you will see a status line reporting
counters only:
`Turn: T  |  Tokens used: U  |  Budget used: B`.
You are responsible for tracking these against the limits above.

Manage your resources strategically:
- Use navigation tools (1 pt) to identify the right papers and
  sections first.
- Use read_section (5 pts) only on sections you have reason to
  believe contain relevant evidence.
- Submit your answer when you have gathered enough evidence. Do
  not exhaust your budget exploring if you already have a strong
  basis for an answer.

{task_block}
\end{verbatim}
\end{tcolorbox}
\label{prompt:agent}

\medskip

\begin{tcolorbox}[
  breakable,
  enhanced jigsaw,
  colback=gray!5,
  colframe=gray!50,
  fonttitle=\bfseries,
  before upper={\small\ttfamily}
]
\captionof{table}{Zero-tool ablation system prompt used in the closed-book replicate (Section~\ref{subsec:ablations} and Appendix~\ref{app:parametric}). All retrieval, navigation, and reasoning tools are removed; the only function call available to the model is \texttt{submit\_answer}.}

\begin{verbatim}
You are a research assistant tasked with answering a multi-hop
scientific question. You must answer based solely on your own
knowledge -- no tools are available.

Read the question and answer options carefully. Use your knowledge
of the scientific literature to reason about which answer is most
likely correct. Consider the claims made in each option, whether the
described findings are consistent with known results in the field,
and the technical plausibility and specificity of each option.

{task_block}

You MUST call submit_answer with your best guess. Reason carefully,
then choose.
\end{verbatim}
\end{tcolorbox}
\label{prompt:zerotool}

\medskip

\paragraph{No-search ablation prompt.}
\label{prompt:nosearch}
The no-search condition reuses Prompt 8 verbatim, but the rendered \texttt{\{tools\_block\}} omits the \texttt{search\_papers} entry; the agent still has access to the six remaining tools and must navigate to evidence through citation traversal alone. No other prompt modifications are made, as specified in Section~\ref{subsec:ablations}.

%% file: sections/Apdx_G.tex

\section{Example Trajectories}
\label{app:example_trajectories}

To make the diagnostic decomposition concrete, we walk through one AgentHop sample (\texttt{mt\_0785}, multi-target depth-2) end-to-end and contrast a successful Claude Opus 4.6 trajectory with a failed Kimi-K2.5 trajectory on the same item.

\subsection{The sample}

\begin{tcolorbox}[
  breakable,
  enhanced jigsaw,
  colback=blue!4,
  colframe=blue!45,
  fonttitle=\bfseries,
  title={Sample \texttt{mt\_0785} (multi-target, depth-2)}
]
\small
\textbf{Seed paper.} \emph{Reflexion: language agents with verbal reinforcement learning} (Shinn et al., 2023; arXiv 2303.11366).

\smallskip
\textbf{Question.} ``Reflexion contrasts itself with earlier LLM-based agents that act in external environments, especially a browser-based question-answering system that depends on human demonstrations/comparisons and a robot planner that grounds candidate actions with affordance scores. Looking across those two foundations, what concrete training or control machinery does each one require, and what headline task outcomes show the kinds of grounded performance Reflexion is trying to replace with verbal reinforcement rather than weight updates?''

\smallskip
\textbf{Options.}
\begin{itemize}\setlength\itemsep{2pt}
  \item[\textbf{A}] (\textsc{NC} -- no\_context distractor) ``WebGPT trains GPT-3 to use a text web browser with behavior cloning followed by reward modelling and RL fine-tuning, and on ELI5 humans preferred WebGPT $\approx$56\% of the time over human-written answers; SayCan multiplies LM skill scores by learned affordance estimates and reaches $\approx$74\% execution success on a real mobile manipulator.''
  \item[\textbf{B}] (\textsc{Gold}) ``WebGPT fine-tunes GPT-3 with behavior cloning, reward modelling, PPO, and best-of-$n$, and its 175B best-of-64 system is preferred 56\% of the time over human demonstrators; SayCan multiplies LM skill probabilities by learned affordance/value estimates, reaching 84\% planning success and 74\% execution success across 101 instructions in the mock kitchen.''
  \item[\textbf{C}] (\textsc{SO} -- seed\_only distractor) ``Earlier agents relied on different training or control machinery, WebGPT used demonstrations/comparisons, SayCan used affordance scores. Reflexion improves over baselines by 22\% on AlfWorld, 20\% on HotPotQA, and 11\% on HumanEval.''
  \item[\textbf{D}] (\textsc{WP} -- wrong\_paper distractor) ``Earlier methods include behavior cloning, reward modelling, RL, and rejection sampling; on toxicity reduction Self-Correct outperforms baselines while on math synthesis a GPT-3 Instruct generator improves from 84.90 to 90.90 on Multitask\dots''
\end{itemize}
\end{tcolorbox}

The gold answer requires synthesizing specific numerical results from \emph{two} different cited papers (WebGPT, arXiv 2112.09332; SayCan, arXiv 2204.01691). Distractor A is plausible-sounding without paper access (parametric reliance), C summarizes the seed paper's own results (navigation failure, did not leave the seed), and D mixes details from a different paper entirely (retrieval error).

\subsection{Successful trajectory: Claude Opus 4.6 (8 turns, 19/30 budget)}

\begin{tcolorbox}[
  breakable,
  enhanced jigsaw,
  colback=blue!4,
  colframe=blue!45,
  fonttitle=\bfseries,
  title={Trajectory 1: Claude Opus 4.6 on \texttt{mt\_0785} (correct, 8 turns, 19/30 budget)},
  before upper={\footnotesize\ttfamily}
]
T0: get\_paper\_info(Reflexion-seed) \\
\quad{}$\to$ ``Title: Reflexion: language agents with verbal reinforcement learning'' \\
\;\;\;\;\;{} get\_references(Reflexion-seed) \\
\quad{}$\to$ ``References for paper $\dots$ (30 papers): WebGPT, SayCan, ReAct, $\dots$'' \\
T1: think \\
T2: list\_sections(WebGPT) \quad\;{}$\to$ 19 sections: A.\,Introduction $\dots$ C.\,Methods, D.\,Evaluation, $\dots$ \\
\;\;\;\;\;{} list\_sections(SayCan)\;\;\,\;{}$\to$ 13 sections: A.\,Introduction $\dots$ C.\,SayCan, E.\,Experimental Evaluation \\
T3: read\_section(WebGPT, "C")  // \emph{Methods} \\
\quad{}$\to$ ``3 Methods. 3.1 Data collection. Guidance from humans is central $\dots$ A language model \\
\quad\quad{}pre-trained on natural language would not be able to use our text-based browser $\dots$'' \\
\;\;\;\;\;{} read\_section(SayCan, "E")  // \emph{Experimental Evaluation} \\
\quad{}$\to$ ``5 Experimental Evaluation. Figure 4: experiments performed in an office kitchen and a \\
\quad\quad{}mock kitchen $\dots$ 5 locations, 15 objects $\dots$ 84\% planning success, 74\% execution success $\dots$'' \\
T4: think \\
T5: read\_section(WebGPT, "D")  // \emph{Evaluation} \\
\quad{}$\to$ ``4 Evaluation. We focused on three WebGPT models, each trained with behavior cloning \\
\quad\quad{}followed by rejection sampling against a reward model $\dots$ best-of-64 $\dots$ 56\% preference rate $\dots$'' \\
T6: think \\
T7: submit\_answer("B")\quad \checkmark\ \textbf{correct}
\end{tcolorbox}

Opus reaches both gold target papers in a single planning step (T0--T2), reads exactly the sections that hold the methodology and headline numbers in parallel (WebGPT Methods + SayCan Experimental Evaluation at T3), thinks, reads WebGPT Evaluation to confirm the 56\% number, thinks again, and submits. Every read is paired with a \texttt{think} call. Budget consumed: 19 of 30 points; turns: 8.

\subsection{Failed trajectory: Kimi-K2.5 (14 turns, 27/30 budget)}

\begin{tcolorbox}[
  breakable,
  enhanced jigsaw,
  colback=blue!4,
  colframe=blue!45,
  fonttitle=\bfseries,
  title={Trajectory 2: Kimi-K2.5 on \texttt{mt\_0785} (incorrect, 14 turns, 27/30 budget)},
  before upper={\footnotesize\ttfamily}
]
T0: get\_paper\_info(Reflexion-seed) \\
T1: get\_references(Reflexion-seed) \\
T2: get\_paper\_info(WebGPT) + get\_paper\_info(SayCan) \\
T3: list\_sections(Reflexion-seed) \\
T4: read\_section(Reflexion-seed, "B")  // \emph{Related work} (\textcolor{red}{wasted}) \\
\quad{}$\to$ ``2 Related work. Reasoning and decision-making. Self-Refine employs an iterative framework $\dots$'' \\
T5: read\_section(Reflexion-seed, "A")  // \emph{Introduction} (\textcolor{red}{wasted}) \\
\quad{}$\to$ ``1 Introduction. Recent works such as ReAct, SayCan, Toolformer, $\dots$'' \\
T6: list\_sections(WebGPT) \\
T8: read\_section(WebGPT, "C")  // \emph{Methods} (gold-paper, gold-section: matches Opus T3) \\
\quad{}$\to$ ``3 Methods. 3.1 Data collection $\dots$'' \\
T9: read\_section(WebGPT, "D")  // \emph{Evaluation} (gold-paper, gold-section) \\
\quad{}$\to$ ``4 Evaluation $\dots$ 56\% preference rate $\dots$'' \\
T10: list\_sections(SayCan) \\
T11: read\_section(SayCan, "C")  // \emph{SayCan} (gold paper but \textcolor{red}{wrong section} -- this is the \\
\qquad\;{} method-description block, not the Experimental Evaluation that contains the 84\%/74\% figures) \\
\quad{}$\to$ ``\textcolor{red}{REJECTED: Insufficient budget (3 remaining, read\_section costs 5).}'' \\
T12: think \\
T13: submit\_answer("A")\quad \ding{55}\ \textbf{wrong} (selected \textsc{NC} distractor)
\end{tcolorbox}

Kimi burns three early turns re-reading the seed paper (T4--T5, sections it already had abstracts for) before pivoting to the gold target papers. It reaches WebGPT successfully but, when it pivots to SayCan, it reads the descriptive ``SayCan'' section (C) rather than the ``Experimental Evaluation'' section (E) that contains the headline 84\%/74\% numbers, and the read is even \emph{rejected} for insufficient budget. With only 3 budget points left and a single \texttt{think} call before submission, Kimi defaults to the plausible-sounding option A (\textsc{NC}). Budget consumed: 27 of 30; turns: 14.

\subsection{What the diagnostic columns record}

For this single sample, AgentHop's diagnostic surface produces:
\begin{itemize}\setlength\itemsep{2pt}
  \item \textbf{Opus:} accuracy $1$, paper recall $1$, section recall $1$ (read both gold sections), conversion $1$, budget $19/30$, turns $8$.
  \item \textbf{Kimi:} accuracy $0$, paper recall $1$ (reached both gold papers), section recall $0$ on SayCan (read ``SayCan'' instead of ``Experimental Evaluation''), conversion contribution $0$, budget $27/30$, turns $14$, NC error.
\end{itemize}
Aggregate accuracy alone marks Kimi wrong without explanation; the decomposition reveals \emph{where} it failed, navigation reached the right papers, but section-level retrieval missed the experimental-evaluation block that contains the gold numbers, and the model defaulted to the parametric distractor under uncertainty.